\documentclass[singlecolumn]{article}
\usepackage{amssymb}
\usepackage{multirow,setspace,verbatim,amsfonts,graphicx,amsmath,amsthm,amsbsy,amssymb,epsfig,url}
\usepackage{amsthm}
\usepackage{gensymb}
\usepackage{booktabs}
\usepackage{cite}
\usepackage{makecell}
\usepackage{epstopdf}

\usepackage{amsfonts}
\usepackage{amsmath,bm}
\usepackage{amssymb}
\usepackage{amsthm}
\usepackage{tikz,graphicx}
\usepackage{mathrsfs} 
\usepackage[algo2e]{algorithm2e} 

\usepackage{algorithm}
\usepackage{array,multirow}
\usepackage{color}
\usepackage{caption}
\usepackage{subcaption}

\newcommand{\Rd}{{\mathbb R}}
\newcommand{\Cd}{{\mathbb C}}

\newcommand{\Sc}{{{\mathcal S}}}

\newcommand{\Tc}{{{\mathcal T}}}

\newcommand{\Yc}{\mathcal{Y}}

\newcommand{\etal}{\emph{et al}}

\newcommand{\Fc}{{\mathcal F}}
\usepackage{amsmath,amssymb,nopageno}
\usepackage[all]{xy}

\theoremstyle{definition} \theoremstyle{plain}

\begin{document}
\begin{titlepage}
\begin{center}
\begin{LARGE}
\begin{bf}

Deep Learning with Domain Adaptation for Accelerated Projection-Reconstruction MR  

\end{bf}
\end{LARGE}
\end{center}

\smallskip

\begin{center}
\begin{large}
{\sc  Yoseob Han$^1$, Jaejun Yoo$^1$, Hak Hee Kim$^2$, Hee Jung Shin$^2$, Kyunghyun Sung$^3$,  and Jong Chul Ye$^1$}
\\[0.1in]
\end{large}

\smallskip

{\em $^1$Dept. of Bio and Brain Engineering\\
Korea Advanced Institute of Science \& Technology (KAIST) \\\
373-1 Guseong-dong  Yuseong-gu, Daejon 305-701, Republic of Korea \\
Email: jong.ye@kaist.ac.kr\\
$^2$ Department of Radiology, Research Institute of Radiology,\\
Asan Medical Center, College of Medicine, University of
Ulsan, Seoul, Korea\\
Email: hhkim@amc.seoul.kr, docshin@amc.seoul.kr\\
$^3$Dept. of Radiological Sciences\\
University of California Los Angeles, Los Angeles, CA, United States\
\\
Email: KSung@mednet.ucla.edu\\
  \vspace*{0.2cm}
  \vspace*{0.2cm}
}
\end{center}

\noindent{ \\
Running Head: Deep Learning for Projection-Reconstruction MR   \\
Journal: Magnetic Resonance in Medicine\\}

\noindent{
$\star$ Correspondence to:\\
Jong Chul Ye,  Ph.D. ~~\\
Professor \\\
Dept. of Bio and Brain Engineering, KAIST \\
373-1 Guseong-dong  Yuseong-gu, Daejon 305-701, Korea \\
Email: jong.ye@kaist.ac.kr \\
Tel: 82-42-350-4320 \\
Fax: 82-42-350-4310 \\
\\
\\
Total word count : approximately 5000 words.

}

\end{titlepage}

\begin{abstract}

\noindent
\textbf{Purpose:} 
The radial k-space trajectory is a well-established sampling trajectory used in conjunction with magnetic resonance imaging. However, the radial k-space trajectory requires a large number of radial lines for high-resolution reconstruction. Increasing the number of radial lines causes longer acquisition time, making it more difficult for routine clinical use.  On the other hand, if we reduce the number of radial lines, streaking artifact patterns are unavoidable. To solve this problem, we propose a novel deep learning approach with domain adaptation to restore high-resolution MR images from under-sampled k-space data.
\\
\textbf{Methods:} 
The proposed deep network removes the streaking artifacts from the artifact corrupted images. To address the situation given the limited available data, we  propose a domain adaptation scheme that employs a pre-trained network using a large number of x-ray computed tomography (CT) or synthesized radial MR datasets, which is then fine-tuned with only a few radial MR datasets.
 \\
\textbf{Results:}
The proposed method outperforms existing compressed sensing algorithms, such as the total variation and PR-FOCUSS methods. In addition, the calculation time is several orders of magnitude faster than the total variation and PR-FOCUSS methods.
Moreover, we found that  pre-training using CT or MR data from  similar organ data is more important than pre-training using data from the same modality for different organ.
\\
\textbf{Conclusion:} 
We demonstrate  the possibility of a domain-adaptation when only a limited amount of MR data is available.
The proposed  method surpasses the existing compressed sensing algorithms in terms of the image quality and computation time. 
\vspace{15 mm}


\end{abstract}
\pagebreak

\newpage

\section*{Introduction}

In MRI, data acquisition is performed by scanning k-space data of objects according to the sampling trajectory. Thus, the quality and the properties of reconstructed images depend on the k-space sampling patterns, such as Cartesian, radial, and spiral sampling patterns \cite{bernstein2004handbook}. 

The radial k-space trajectory has many advantages over the Cartesian k-space trajectory, although the Cartesian k-space trajectory is the most widely used sampling pattern. Specifically, each line measured using the radial k-space trajectory contains both center and periphery frequency information. Since the center frequencies are over-sampled for all data lines, the center position is averaged when image reconstruction is conducted such that motion artifacts from the flow or respiration are reduced with the radial k-space trajectory \cite{jung1991reduction, gmitro1993use, katoh2006mr, trouard1996analysis}. In addition, the radial k-space trajectory for sub-sampling experiments shows visually better image quality than the under-sampled Cartesian k-space trajectory, as the associated aliasing artifacts appear as less-coherent streaking patterns that are much less disturbing than wrap-around artifact patterns due to the under-sampling in the Cartesian k-space.

Because the increased scan time is a critical weak point, many investigators have employed compressed sensing  \cite{donoho2006compressed, candes2006robust,lustig2007sparse,jung2009k} in an effort to reduce the scan time.
In fact, based on the observation that streaking artifacts are less coherent,  the radial trajectory  was the first sampling pattern used for the ground-breaking  compressed sensing (CS) work by Romberg and Candes \cite{candes2006robust},
and many researchers have developed under-sampled radial k-space image reconstruction methods using compressed sensing (CS) algorithms \cite{ye2007projection,jung2010radial,feng2014golden,cruz2015accelerated}.  One of the remarkable
aspects of the CS approaches is  that  although the corrupted image appears to lose fine detail due to the down-sampling, the seemingly lost information can be recovered by iteratively imposing sparsity in the total variation, wavelets, and other transform domains. 
%
%
%
However, one of the main technical limitations of CS methods is the increased computational complexity due to the iterative reconstruction process, thus necessitating a new method that obviates these limitations.

Recently, deep learning approaches have achieved tremendous success in various fields, such as classification \cite{krizhevsky2012imagenet}, segmentation \cite{ronneberger2015u}, denoising \cite{zhang2016beyond}, and super resolution \cite{kim2015accurate, shi2016real}. In the field of medical imaging, most works have focused on image-based diagnostics. 
Recently, Wang \etal.\cite{wang2016accelerating} applied deep learning to CS-MRI. They trained a deep neural network from downsampled reconstruction images to learn an instance of fully sampled reconstruction. 
Subsequently, they used the deep learning result either for initialization or as a regularization term in classical CS approaches. 
A deep network architecture using an unfolded iterative CS algorithm has also been proposed \cite{hammernik2016learning}. 
Instead of using handcrafted regularizers, the authors attempted to learn a set of optimal regularizers. 
A multilayer perceptron was also proposed for accelerated parallel MRI \cite{kwon2016learning}.
In the x-ray computed tomography (CT) area, Kang \etal.\cite{kang2016deep} provided the first systematic study of a deep CNN for low-dose CT and showed that a deep CNN using directional wavelets can more efficiently remove low-dose-related CT noises. Unlike these low-dose artifacts from reduced tube currents, the streaking artifacts originating from sparse projection views show globalized artifacts that are difficult to remove using conventional denoising CNNs \cite{chen2015learning, mao2016image, xie2012image}. Jin \etal.\cite{jin2016deep} and Han \etal.\cite{han2016deep} independently proposed multi-scale residual learning networks using U-Net \cite{ronneberger2015u} to remove the global streaking artifacts caused by sparse projection views.

Recall that the projection slice theorem can convert the radial k-space data into a sinogram data format corresponding to parallel beam x-ray CT \cite{hsieh2009computed}.
Accordingly,  the radial k-space sampling data can be reconstructed by a CT reconstruction method, such as filtered back-projection (FBP).
This does not require re-gridding of the k-space data to the Cartesian grid, which makes the reconstruction more accurate. This method is often referred to as projection reconstruction (PR) in the MR literature. The degree of similarity between sparse-view CT and accelerated radial acquisition in MR allows us to exploit the synergy from deep learning-based CT reconstruction  \cite{han2016deep}. 

In particular, given this similarity between projection-reconstruction MR and CT, this paper proposes a novel means of {\em domain adaptation}  \cite{yosinski2014transferable, pan2010survey} of a type never been exploited in MR reconstruction
approaches which rely on deep learning \cite{wang2016accelerating,hammernik2016learning}.
Note that a large dataset is usually required when training a deep neural network. In contrast to x-ray CT, the radial trajectory is not widely used in MRI; hence, a large number of PR datasets cannot be easily obtained in most MR sites. 
Thus, we want to transfer an advanced deep network from a CT dataset and  adapt the network parameter to suit MR reconstruction.
 Although our deep network could be trained with only radial MR data, one of the most important innovations and focuses of this work is  to consider this common scenario using domain adaptation, which is an active field of research in deep learning  \cite{yosinski2014transferable, pan2010survey}.
Given a pre-trained network from CT, the actual MR training only requires very few radial MR datasets, which significantly reduces the training time and expands the applicability of deep learning for MR imaging.
By extending this idea furthermore,  we also demonstrated that  pre-training  using synthetic radial MR data from public domain 
MR images 
is also an effective approach for domain adaptation when the underlying structure are similar.
This finding suggests that the domain adaptation can be widely used to supplement the limited training data set in machine learning approach for MR reconstruction.

Note that our approach can be more accurately classified as a restoration method for artifact removal rather than as a reconstruction method because the trained network is used to restore the image from an image contaminated with streaking artifacts. However, the final goal is to restore high-resolution images from insufficient data, as it is in deep learning-based reconstruction methods \cite{wang2016accelerating,hammernik2016learning}.


\section*{Theory}

\subsection*{MR Radial k-space Trajectory versus CT Sinogram}

If $m(x,y)$ denotes an MR image,  then the radial k-space data  obtained at the angle $\theta$ can be expressed as
\begin{eqnarray}\label{eq:mr_sampling}
\hat P_\theta(\omega) 
 &=& \int\int dx dy {~m(x,y)e^{-j\omega( x\cos(\theta)+y\sin(\theta))}}.
 \end{eqnarray}
Because this is a Fourier transform in the polar coordinates, the inverse transform is given by :
\begin{eqnarray}\label{eq:fbp}
m(x,y)=\frac{1}{2\pi} \int_0^{\pi}d\theta\int_{-\infty}^\infty |\omega| \hat P_\theta (\omega) e^{j\omega} d\omega \ .
\end{eqnarray}
 which is often referred to as filtered back-projection (FBP). 
In this equation, $|\omega|$ denotes a ramp filter. 
 In fact, the FBP formula in \eqref{eq:fbp} is the standard reconstruction formula for parallel-beam X-ray CT, where $\hat P_\theta(\omega)$ is obtained by taking the 1-D Fourier transform along the $t$ direction of the projection data $P_\theta(t)$.
 More specifically, we have
 the projection slice theorem \cite{hsieh2009computed}:
 \begin{eqnarray*}
\hat P_\theta(\omega)  
  &=& \int dt e^{-j\omega t} \int\int dx dy {~m(x,y)\delta(t-x\cos\theta-y\sin\theta)} \nonumber \\
    &=& \int dt P_\theta(t)   e^{-j\omega  t} \nonumber \\
    &=& \mathcal{F}\{P_\theta(t) \} 							
\end{eqnarray*}
where $P_\theta(t):=\int\int dx dy {~m(x,y)\delta(t-x\cos\theta-y\sin\theta)} $ is the projection data at angle $\theta$.
This similarity between radial acquisition and CT implies that a deep network trained from CT data can be used synergistically in projection-reconstruction MRI.
Since the formulations in \eqref{eq:mr_sampling} and \eqref{eq:fbp} are continuous-domain formulation, in practice the
k-space data $\hat P_\theta(\omega)$ are sampled along angle direction and read-out direction, forming a discrete matrix:
$$\hat P_\theta(\omega) \longrightarrow \hat P \in \Cd^{N_\theta\times N_\omega}.$$
Similarly, the image $m(x,y)$ is also reconstructed on a discrete grid:
$$m(x,y)  \longrightarrow  M \in \Cd^{N_x \times N_y}.$$

\begin{figure}[!b] 	
\centering
{\includegraphics[width=1.0\linewidth]{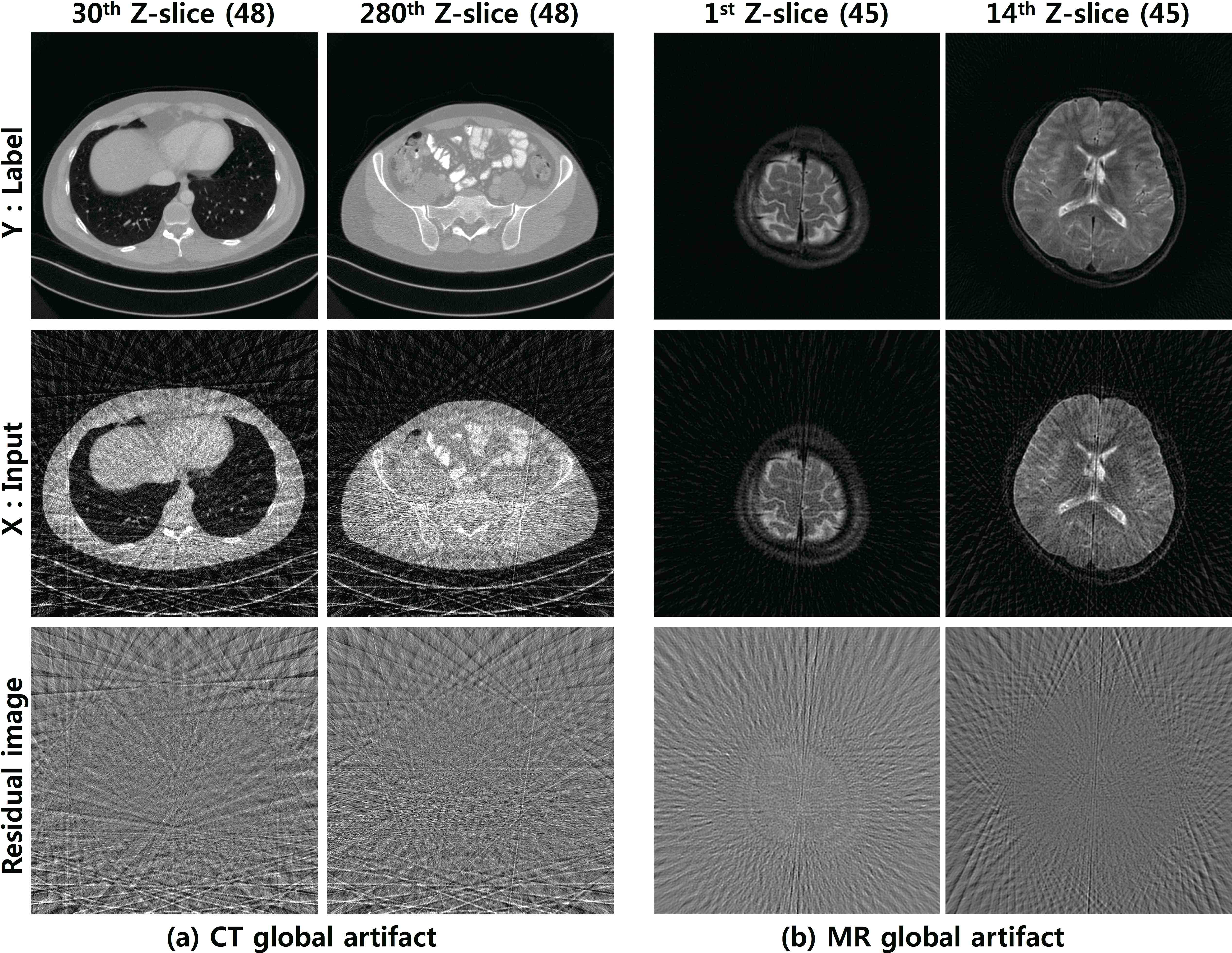}}
\caption{Several global artifact patterns for (a) reconstruction images from 48 view projections in CT and (b) reconstruction images from 45 radial k-space trajectory in MR.}
\label{fig:streaking_artifact}
\end{figure}

With any insufficiency with regard to the number of radial k-space scan lines $N_\theta$,  streaking artifacts occur in the reconstructed image from MR and CT. For example, 
Figs. \ref{fig:streaking_artifact}(a) and (b) show the FBP reconstruction results of CT and MR from 48 view projection and 45 radial k-space lines, respectively. Here, the streaking artifacts were obtained from the magnitude images.
In Fig. \ref{fig:streaking_artifact}(a),  the first  and the second columns show very different CT images, but similar streaking artifact patterns are observed from those images. Interestingly,  similar phenomena were also observed in the MR images, as shown in Fig. \ref{fig:streaking_artifact}(b). 
Thus, we are interested in transferring a network trained from CT domain for MR image restoration. This can
be done using domain adaptation which is the main topic in the following section.

\subsection*{Domain Adaptation}

A deep neural network typically requires a large number of datasets. Unfortunately, a Cartesian k-space sampling pattern is much more widely used in MRI than a radial sampling pattern, and for this reason it is often difficult to collect many datasets for radial sampling patterns. On the basis of the observation that streaking artifact patterns are similar regardless of the imaging modalities, we utilize a domain-adaptation technique to overcome the insufficiency of radial MR dataset. 

Specifically, 
$x\in X\subset \Rd^{N_x\times N_y}$ denotes the real-valued {\em magnitude} image  of the sparse-view reconstruction,
and $y \in Y\subset \Rd^{N_x\times N_y}$ corresponds to the streaking artifact-free images in the {\em magnitude}  domain.
We define a {\em domain} as a pair consisting of a distribution $D$ over $X$ and a regression function
 $f:X\mapsto Y$ that belongs to the function class $\Fc$ as represented by a neural network.
We consider two domains: a {\em source} domain and a {\em target} domain. In this paper, the source domain is CT data or synthesized radial MR data, 
while the target domain is  in vivo radial MR data.

We denote by $Q$ the source domain and by $P$ the target domain distributions.
 In the domain-adaption scenario, we receive two samples: a labeled 
 sample of $m$ points from the source domain  $\Sc =\{ (x_1,y_1),\cdots, (x_m, y_m)\}$ with
 $x_1,\cdots, x_m \in \Rd^{N_x\times N_y}$ drawn according to $Q$, matched with the target sample $y_1,\cdots, y_m \in \Rd^{N_x\times N_y}$, 
 and an unlabeled sample drawn  according to the target distribution $P$.
In addition, a small amount of label data from the target domain $\Tc'=\{(x_1',y_1'),\cdots, (x_s',y_s')\}$ are available.
This is based on the practical acquisition scenario in which there are ample projection datasets from CT or synthesized radial MR, whereas relatively few in vivo radial MR datasets are available.
 
 For any distribution $D$, a loss function is defined as a function that minimizes the risk in the target domain, i.e., $$L_D(y,f) =  E_{x\sim D} \left\|y-f(x;\eta)\right\|^2,$$ 
where $\eta\in \Rd^{N_\eta}$ denotes the network parameters, and $x, y\in \Rd^{N_x\times N_y}.$
Hence, our goal in the domain adaptation is to minimize the loss in the target (i.e., MR) domain.
By means of triangular inequality, we have the equations below
\begin{eqnarray*}
L_P(y,f) & \leq  & L_Q(y,f) +  \mathrm{disc}_\Yc(P,Q),
\end{eqnarray*}
where what is termed the $\Yc$-discrepancy between two domains is defined by \cite{ben2007analysis, ben2010theory} 
 \begin{equation}
 \mathrm{disc}_\Yc(P,Q) = \sup_{f\in \Fc}\left| L_P(y,f)-L_Q(y,f) \right|.
 \end{equation}
To minimize the risk in the target domain, we are interested  in minimizing the
 risk in the source domain $L_Q(y,f)$ and the  $\Yc$-discrepancy between the two domains.
Note that all the calculation of $L_Q(y,f)$ and $\Yc$-discrepancy is in the magnitude domain, so we do not need to consider
the complex nature of the MR image.

However, there are several technical issues. The first of these is that the associated probability distribution $Q$ is unknown.
To address this issue, approaches from  statistical learning theory  \cite{anthony2009neural} 
offer the bound of  the risk of a learning algorithm with regard to a complexity measure (e.g., VC dimension, shatter coefficients) and empirical risk. 
Specifically, with a probability of $\geq 1-\delta$ for some small $\delta>0$,  for every  function $f\in \Fc$,  we have \cite{bartlett2002rademacher}
\begin{equation}\label{eq:L}
L_Q(y,f) \leq \underbrace{ L_{\hat Q}(y,f)}_{\text{empirical risk}} + \underbrace{2 \hat R_m(\Fc)}_{\text{complexity penalty}}+ 3 \sqrt{\frac{\ln(2/\delta)}{m}},
\end{equation}
where the empirical risk is given by
$$ L_{\hat Q}(f) = \frac{1}{m}\sum_{i=1}^m \|y_i - f(x_i;\eta)\|^2$$
and the Rademacher complexity $\hat R_n(\Fc)$ is defined by
$$\hat R_m(\Fc)=  E_\sigma \left[\sup_{f\in \Fc} \left(\frac{1}{m}\sum_{i=1}^m \sigma_i f(x_i) \right) \right],$$ 
where $\sigma_1,\cdots, \sigma_n$ are independent random variables uniformly chosen  from $\{-1,1\}$.
Thus, we can obtain the following generalization error bound for the domain adaptation:
\begin{eqnarray*}
L_P(y,f) & \leq  &{ L_{\hat Q}(y,f)} +  \mathrm{disc}_\Yc(P,Q) +  {2 \hat R_m(\Fc)}+ 3 \sqrt{\frac{\ln(2/\delta)}{m}}
\end{eqnarray*}

In general, the Rademacher complexity term is determined by the structure of the network. In order to reduce the generalization error of  domain adaptation for MR reconstruction, we should therefore minimize the empirical risk in the source domain and the discrepancy between the two domains.
Empirical risk minimization in the source domain is a standard neural network training process. In our case, we use CT data or synthesized radial MR data so that the network can learn the mapping from artifact-corrupted image to artifact-free image. To minimize discrepancies, we use a small set of label data from radial  MR data $\Tc'=\{(x_1',y_1'),\cdots, (x_s',y_s')\}$ with $x_i',y_i'\in \Rd^{N_x\times N_y}$ so that the refined network output follows the target distribution.
In particular, we use an incremental approach, in which the network parameter is iteratively adjusted to change the network outputs $f(x_s')$ to reduce the discrepancy.

Specifically, we denote  the estimated network parameter at the $t$-th iteration of our algorithm $\eta^t\in \Rd^{N_\eta}$: each iteration of
our method takes a batch of random input $\{x_i'\}$ and calculates its corresponding
output $\hat y_i' = f(x_i';\eta^t)$ based on $\eta^t$.
To realize a better approximation of the target distribution $P$, we restrict the path of evolution
such that 
\begin{eqnarray}\label{eq:Delta}
y_i'' = \hat y_i'+\epsilon \Delta y_i'&,& \mbox{where}\quad
\Delta y_i'= y_i'-\hat y_i' =  y_i'- f(x_i';\eta^t) \in \Rd^{N_x\times N_y}.
\end{eqnarray}
Therefore, we should update the network parameter such that
$$\eta^{t+1} \leftarrow \arg\min_{\eta} \sum_{i=1}^b \|f(x_i';\eta)-y_i''\|^2,$$
where $b$ denotes the batch size.
Using the Taylor series expansion, we have
\begin{eqnarray*}
 \|f(x_i';\eta)-y_i''\|^2 \simeq  \|\partial_\eta f(x_i';\eta^t)(\eta-\eta^t)-\epsilon \Delta y_i'\|^2
\end{eqnarray*}
Accordingly, we have
\begin{eqnarray}\label{eq:eta}
\eta^{t+1} = \eta^t+\epsilon \Delta \eta^t, & \mbox{where} & \Delta\eta^t= \arg\min_\delta  \sum_{i=1}^b\|\partial_\eta f(x_i';\eta^t)\delta-\epsilon \Delta y_i'\|^2.
\end{eqnarray}
The one-step  stochastic gradient descent of \eqref{eq:eta} is then given by
\begin{eqnarray}
\eta^{t+1}  &=&  \eta^t+\epsilon \sum_{i=1}^b \partial_\eta f(x_i';\eta^t)\Delta y_i'  \notag \\
&=&\eta^t+\epsilon \sum_{i=1}^b \partial_\eta f(x_i';\eta^t) \left(y_i'- f(x_i';\eta^t) \right)  \label{eq:update},
\end{eqnarray}
where $y_i'\in \Rd^{N_x\times N_y}$ denotes the MR artifact-free image in the magnitude domain and $ f(x_i';\eta^t)$ is the $t$-th iteration network output for a given MR input data $x_i'\in \Rd^{N_x\times N_y}$.
This is then back-propagated across all network levels to update the network parameters. 

Note that there are different approaches for domain adaptation, such as fine-tuning \cite{long2015fully} and adversary domain adaptation \cite{ganin2015domain}.
The difference in these approaches is essentially the definition  of $\Delta y_i'$ in \eqref{eq:Delta} and \eqref{eq:update}.
The aforementioned approach using \eqref{eq:Delta} corresponds to network fine-tuning using the target domain data, but the algorithm can be improved using more advanced domain-adaptation techniques. 

%
%
%

\subsection*{Network Architecture}


As shown in Fig.~\ref{fig:streaking_artifact}, the streaking artifacts are displayed as globally distributed artifacts.
In order to capture globally spread streaking artifacts, the receptive field of the network should be sufficiently large and
the network depth should be increased. Thus, multi-scale learning network using U-net  \cite{ronneberger2015u}  with residual path is useful
due to its large receptive field.

\begin{figure}[!b] 	
\centering
{\includegraphics[width=1.0\linewidth]{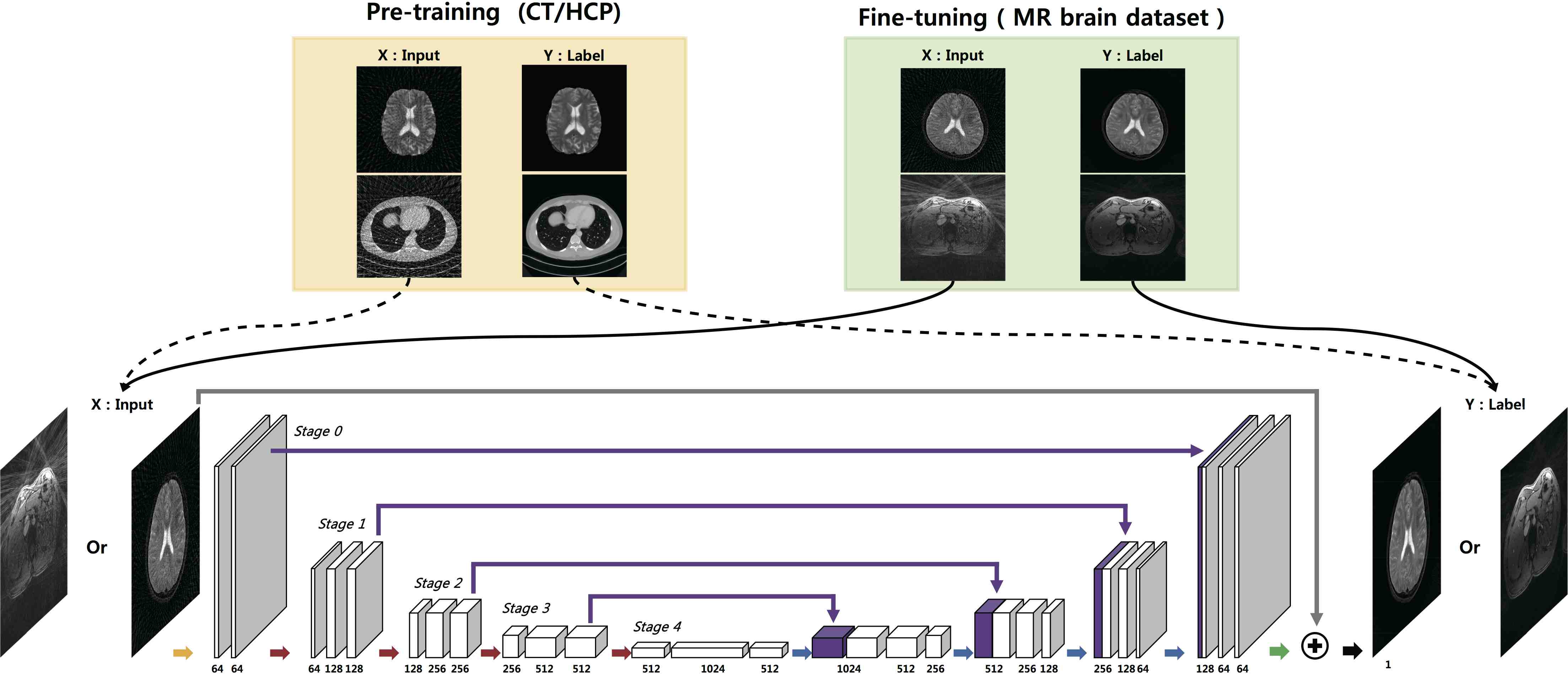}}
\caption{The proposed domain adaptation architecture for radial k-space under-sampled MR. }
\label{fig:transfer_learning}
\end{figure}

Specifically,  as illustrated in Fig \ref{fig:transfer_learning}, in the first half of the proposed network, each stage is followed by a max pooling layer, whereas in the latter half of the network, an average unpooling layer is used.
In our previous work \cite{han2016deep}, we showed  that this results in an enlarged effective receptive field, which is more advantageous when removing streaking artifacts.
In addition, scale-by-scale contracting paths are used to concatenate the results from the front part of the network to the latter part of the network.  The numbers of channels for each of the convolution layers are shown in Fig \ref{fig:transfer_learning}. Note that this number is doubled after each pooling layer. 
In our network, zero-padding was used for the convolution so that the image size does not decrease while passing through the convolution layer.  

In addition, the proposed residual network consists of a convolution layer, batch normalization \cite{ioffe2015batch}, a rectified linear unit (ReLU) \cite{krizhevsky2012imagenet}, and the contracting path connection with concatenation \cite{ronneberger2015u}. Specifically, each stage contains four sequential layers composed of convolution with $3 \times 3$ kernels, batch normalization, and ReLU layers. The last stage has two sequential layers, and the last layer contains only a convolution layer with a $1 \times 1$ kernel. 
Finally, the proposed U-net has residual path inspired by the Residual Net \cite{he2016deep}.
This additional residual path facilitates the training by providing additional path for the back-propagation \cite{he2016identity}.


\subsection*{Data Type}


In neural network training of MR images, 
care should be taken  owing to the mismatch in the data type.
Note that the input and output data   in our pre-trained network are the real data type, since projection data from CT is real-valued.
The situation is the same when the projection data is synthesized from real-valued MR public data set.
On the other hand,
 the data obtained from a real MR system is a complex data type. Therefore, we need to convert the MR data to the real-valued image.
 
Specifically, for the projection-reconstruction MRI, there are two types of back projection methods: magnitude backprojection and complex backprojection \cite{peters2000undersampled}.
In both methods, to perform back-projection, the complex radial k-space data is first zero-padded 
along the radial direction, after which 1-D FFT is conducted to generate complex-valued sinogram data.
For magnitude back-projection, we then compute the magnitude of the sinogram data and perform FBP.  
In complex back-projection, the real and imaginary channels are filtered and back-projected separately. 
For  parallel imaging using multiple coils, the magnitude or complex back-projection is performed separately for each coil data.
Then, we calculate the square root of the square-sum-of-square (SSOS) image to combine the complex-valued and/or multi-coil images. 
Thus, the final reconstructed image from both magnitude and complex backprojection
becomes a real-valued image, and our network fine-tuning is performed to learn the real-valued artifact-free magnitude images
 from a real-valued artifact-corrupted image in the magnitude domain.
 Thus, we do not need to be concerned about the data-type mismatch between the pre-training and fine-tuning.


\section*{Materials and Methods}

\subsection*{Data Acquisition}


Recall that the proposed method consists of two parts: empirical risk minimization using pre-training in the source domain and the 
discrepancy minimization  by means of fine-turning in the target domain. 
Thus, we need two data sets: one for the source domain and the other for target domain.

For pre-training with CT data, rebinned fan-beam CT data  in our previous study \cite{han2016deep} were used to generate the artifact-corrupted
and artifact-free images.  The CT data was from 3602 slices of actual patient CT data $512 \times 512$ in size provided by the 2016 AAPM low-dose CT Grand Challenge  (http://www.aapm.org/GrandChallenge/LowDoseCT/). 
Specifically, the data were acquired from Somatom Definition AS+ and Somatom Definition Flash (manufactured by Siemens Healthcare, Germany) operated in single-source mode. The X-ray tube potential and mAs value were 120 kVp and 200 mAs, respectively. Here, the flying focal spot (FFS) technique \cite{flohr2005image, kachelriess2006flying} served as the means of data acquisition. Because a rebinning process is required to reconstruct images with FFS data, the FFS data was rebinned into fan-beam CT data, and fan-beam FBP is used to obtain the artifact-free images and artifact-corrupted images from full-views and sparse-view data.

In addition,  another pre-training  were performed using a synthetic radial data from public MR data set. To generate the synthetic  radial data, T2 weighted images from 
Human Connectome Project (HCP) MR dataset   (https://db.humanconnectome.org) were used. The T2 weighted brain images contained within the HCP were acquired by Siemens 3T MR system using a 3D spine-echo sequence. The repetition time (TR) and echo time (TE) were 3200 ms and 565 ms, respectively. The number of coils was 32. The field of view (FOV) defined $224 \times 224 ~\rm{mm}^2$ and the voxel size was 0.7 mm. 
From the HCP data, the parallel beam projection data are obtained using radon() function in MATLAB.

Once a pre-trained network is given, fine-tuning can be performed using only a few in vivo radial MR datasets  to minimize the discrepancy between the source and target domains. 
In our work, fine-tuning was performed for brain and abdomen MR datasets separately using both pre-trained networks from CT and HCP data set.
Here, we aim to investigate the effect of imaging modality (CT vs. MR)  and organ (brain vs. abdominal area) in the domain adaptation.

\begin{figure}[!b] 	
\centering
{\includegraphics[width=0.9\linewidth]{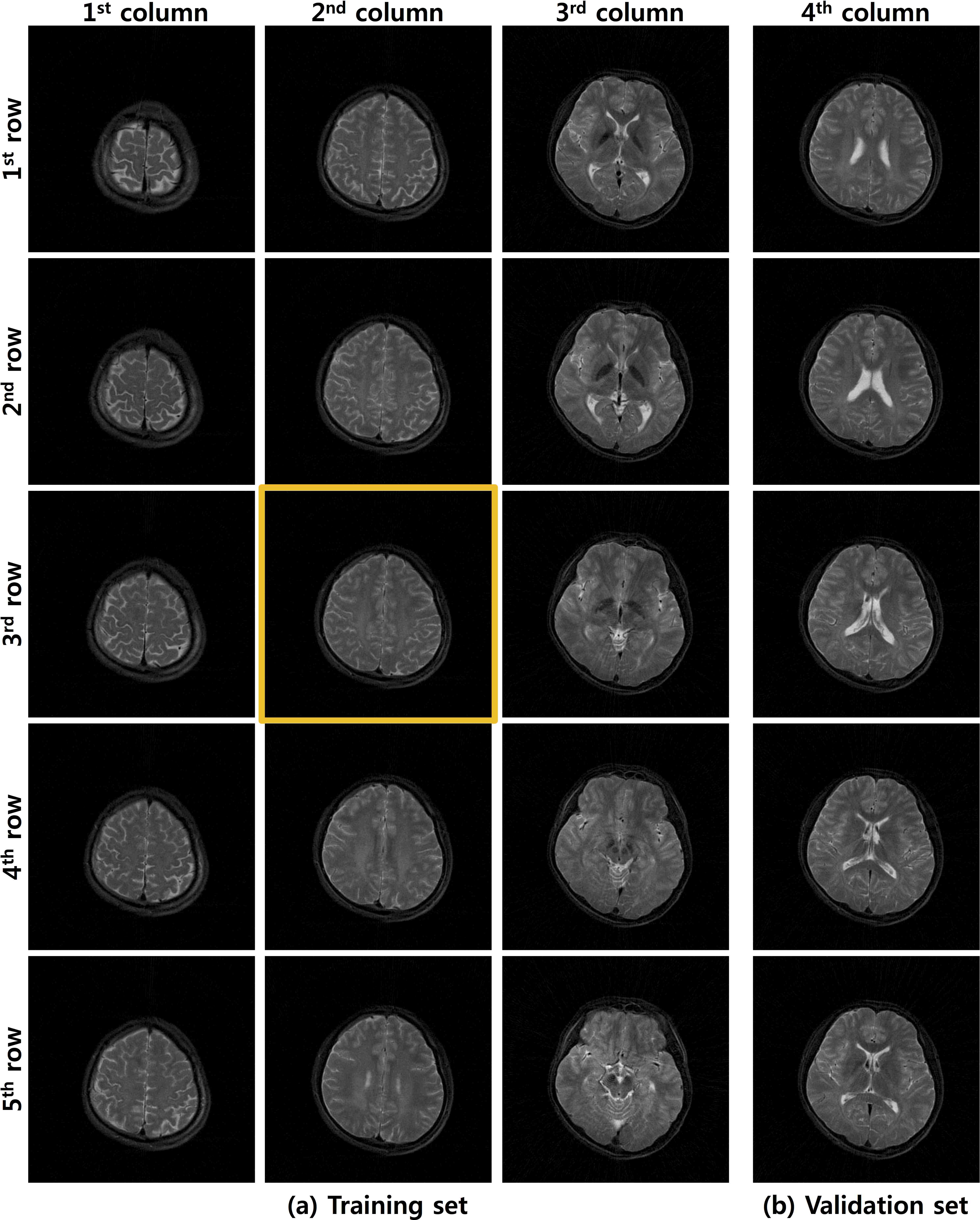}}
\caption{The training and validation dataset acquired from subject 0. In (a), 1, 3, 6, 9, and 15 slices from 15 slice images were used in the training dataset. Validation was performed using five distinct slices, as shown in (b). In order to observe the effect of the number of MR dataset, the network parameter was fine-tuned with various numbers of training datasets composed of  1, 3, 6, 9, and 15 MR slices. If only one MR slice was used, the yellow box image in (a) was used. If 3,  6 or 9 slices were used for domain adaptation, then the dataset  in the $(3)$, $(2, 4)$, or $(1, 3, 5)$ rows of (a) were used. When using 15 slices in the training dataset, all of the dataset in (a) were used.
}
\label{fig:mr_dataset}
\end{figure}

Specifically, the real radial MR data of brain were acquired from a radial spin-echo sequence by a 3.0T MR system manufactured by ISOL technology in Korea. The TR and TE were 2000 ms and 90 ms, respectively. There were 20 slices in total, and the thickness of each slice was 4 mm. The FOV was $220 \times 220~\rm{mm}^2$, and one coil was used. In total, 256 points k-space data instances were obtained for each projection. Fully sampled projection MR data with 180 projection views were acquired from subject 0 for training and validation.
Fig. \ref{fig:mr_dataset} shows the brain MR dataset collected from subject 0 for both training and validation. As shown in Fig. \ref{fig:mr_dataset}(a), 1, 3, 6, 9, and 15 slices from 15 slice images were used in the training dataset. Validation was performed using five distinct slices, as shown in Fig. \ref{fig:mr_dataset}(b). In order to observe the effect of the number of MR datasets, the network parameter was fine-tuned with various numbers of training datasets composed of  1, 3, 6, 9, and 15 MR slices.
If only one MR slice was used, the yellow box image in Fig. \ref{fig:mr_dataset}(a) was used.
 If 3,  6 or 9 slices were used for domain adaptation, then the datasets in rows $(3)$, $(2, 4)$, and $(1, 3, 5)$ in Fig. \ref{fig:mr_dataset}(a) were used.
When using 15 slices in the training datasets, all of the datasets in Fig. \ref{fig:mr_dataset}(a) were used.
In addition, fully sampled and in vivo accelerated radial MR data consisting of 180, 90, and 45 projection views were obtained from subjects 1 and 2. This dataset was used during the test phase to evaluate the generalization performance of the trained network.

In addition, liver MRI scans of seven healthy volunteers were acquired on the axial plane using a 3D stack-of-radial sequence with golden angle ordering. 
The experiments were performed on a 3T MRI system (PrismaFit, Siemens Healthcare GmbH, Erlangen, Germany) 
after the subjects provided informed consent according to a protocol approved by the local Institutional Review Board (IRB) of UCLA.
Each subject was instructed to take a 19 second breath-hold, and the acquisition was performed during the breath-hold.
The acquisition parameters were as follows: flip angle = 10$^{\circ}$, TR = 2.35 ms, TE = 1.12 ms, and FOV = $340 \times 340~\rm{mm}^2$. There were 20 coils.
Due to the golden angle ordering, we retrospectively under-sampled k-space data at various acceleration factors 
by simply taking a portion of the fully sampled radial spokes (302 projections views and 192 points per projection),
and we removed the first and last slices out of 20 slices due to artifacts in the fully sampled data.
Similar to the brain MRI data, subject 0 was used for training and validation
with 15 and 3 slices as the training and validation datasets, respectively. 
The data for the remaining patients were used during the test phase.

 The magnitude and complex back-projection were used to obtain both under-sampled and fully-sampled reconstruction images in the brain and abdominal dataset, respectively.
Subsequently, we conducted experiments with several downsampled factors to learn the  streaking artifact from various downsampling factors,  where the artifact-contaminated images from all downsampling factors were used for training. 
To perform the back-projections, the complex radial k-space data was first zero-padded with 729 samples along the radial direction, after which 1-D FFT was performed to generate complex-valued
sinogram data.
 The resulting size of the reconstructed real-valued images was $512 \times 512$ for both the brain and abdomen images.
 As described before, the images are finally obtained as the magnitude image from complex-valued reconstruction, and for parallel imaging,
 the images are obtained as the square root of the sum of square (SSOS) image by combining the complex-valued and/or multi-coil images.

Here, it is important to note that one of the most important advantages of domain adaptation is that we do not need additional pre-training for our MR reconstruction problem if there exists  an available pre-trained network \cite{han2016deep}. Hence, the computational cost of pre-training for the domain-adaptation scheme can be often considered as zero.

\subsection*{Network Training}

The pre-trained network from CT data was trained by means of the stochastic gradient descent (SGD) method. The regularization parameter was $\lambda = 10^{−4}$. The learning rate ranged from $10^{-3}$ to $10^{-5}$, and was gradually reduced at each epoch.

Another pre-training was also performed with the synthetic radial MR dataset.  T2 weighted magnitude images of HCP dataset were used as artifact-free images, from which synthetic projection MR data were generated  using the $radon$ operator in MATLAB. The input images with global artifact caused by insufficient radial acquisition were re-generated with $iradon$ operator using 36, 45, 60, and 90 views projection MR data. As a training dataset, we used 100 subject data, including 3600 slices images.
The network parameters were the same as those for the CT pre-trained network.


For given pre-trained networks, fine-tuning was then performed using a small number of MR slices to adjust for the difference between the two domains.
The fine-tuning procedure was identical to that in pre-training with a stochastic gradient algorithm using the same parameters except the learning rate. Specifically, when CT pre-trained network was used, the learning rate is applied to $10^{-4}$ to $10^{-6}$. On the other hand, the learning rate is $10^{-5}$ to $10^{-7}$ if synthetic
radial MR  pre-trained network was used. Other slices with very different structures were then used as the test set for validation. 
We used 1000 and 3000 epochs for the brain data and the abdominal data, respectively, including the pre-training. Mini-batch data was used using an image patch, and the size of image patch was $256 \times 256$.  
Since CNN learns the convolutional filters which are spatially invariant, the same filter can be used
for reconstructing $512\times 512$ images. 
Using the $256 \times 256$ patch size significantly reduces network training time.

The proposed network is designed for reconstruction from 2, 3, 4, and 5 accelerations. To learn various acceleration factors simultaneously, the network was trained using several down-sampling images together. Specifically,  for the brain dataset,  reconstructions from 180 views were considered as artifact-free data; consequently, the input images used to train the brain dataset consist of batches of reconstructed images using 90, 60, 45, and 36 views. For the abdominal data, 302 views were used to reconstruct the artifact-free image. Thus, the training dataset is composed of downsampled images with 151, 100, 75, and 60 views. The simultaneous training using multiple acceleration factors was helpful in that it reduced over-fitting during the network training process.

The network was implemented using the MatConvNet toolbox (ver.20) \cite{vedaldi2015matconvnet} in the MATLAB 2015a environment (MathWorks, Natick). We used an NVidia GTX 1080 Ti graphic processor and an Intel i7-7770 CPU (3.60GHz).

\subsection*{Performance Evaluation}

 In order to  demonstrate the comparative advantage of our method,   we have compared with CS algorithms including total variation (TV) \cite{goldstein2009split}-based CS \cite{candes2006robust} and PR-FOCUSS \cite{ye2007projection}. The TV result was calculated by solving the following TV minimization equation,
$$\min_{x} \frac{1}{2} ||y - Ax||^2_2 + \lambda TV(x),$$ 
where $A$ is the subsampled projection operator, $y$ is the sub-sampled measurement,  $TV(x)$ denotes the TV penalty,  and $\lambda$ is the regularization parameter.
The TV minimization problem was solved using the alternating direction method of multiplier (ADMM) algorithm with variable splitting with a regularization parameter of $\lambda= 10^{-2}$ and an iteration number of 100.
The PR-FOCUSS reconstruction scheme utilized the parameter settings from our earlier work \cite{ye2007projection}.

 For a quantitative evaluation, we used the normalized mean square error (NMSE).
  The NMSE is defined by
\begin{eqnarray*}
NMSE = \frac{||x - \hat x||^2_2}{||x||^2_2},
\end{eqnarray*}
where $x$ denotes the reconstruction results from full views and  $\hat x$ is the reconstruction results from the sub-sampled projection data. The average NMES values were calculated by averaging all slice reconstruction results.

 \section*{Results}\label{sec:result}

To evaluate the effective domain adaptation schemes, we have evaluated two pre-trained networks: one from CT data and the other from  synthetic radial MR 
data from HCP data set.  Note that the CT data from AAPM low-dose CT challenges are mainly from abdominal imaging, whereas
the HCP data are for brain imaging. Thus, the source and target domains are different in their imaging modalities and target organs, so we are interested in investigating important factors for domain adaptation.

\begin{figure}[!b] 	
\centering
{\includegraphics[width=0.8\linewidth]{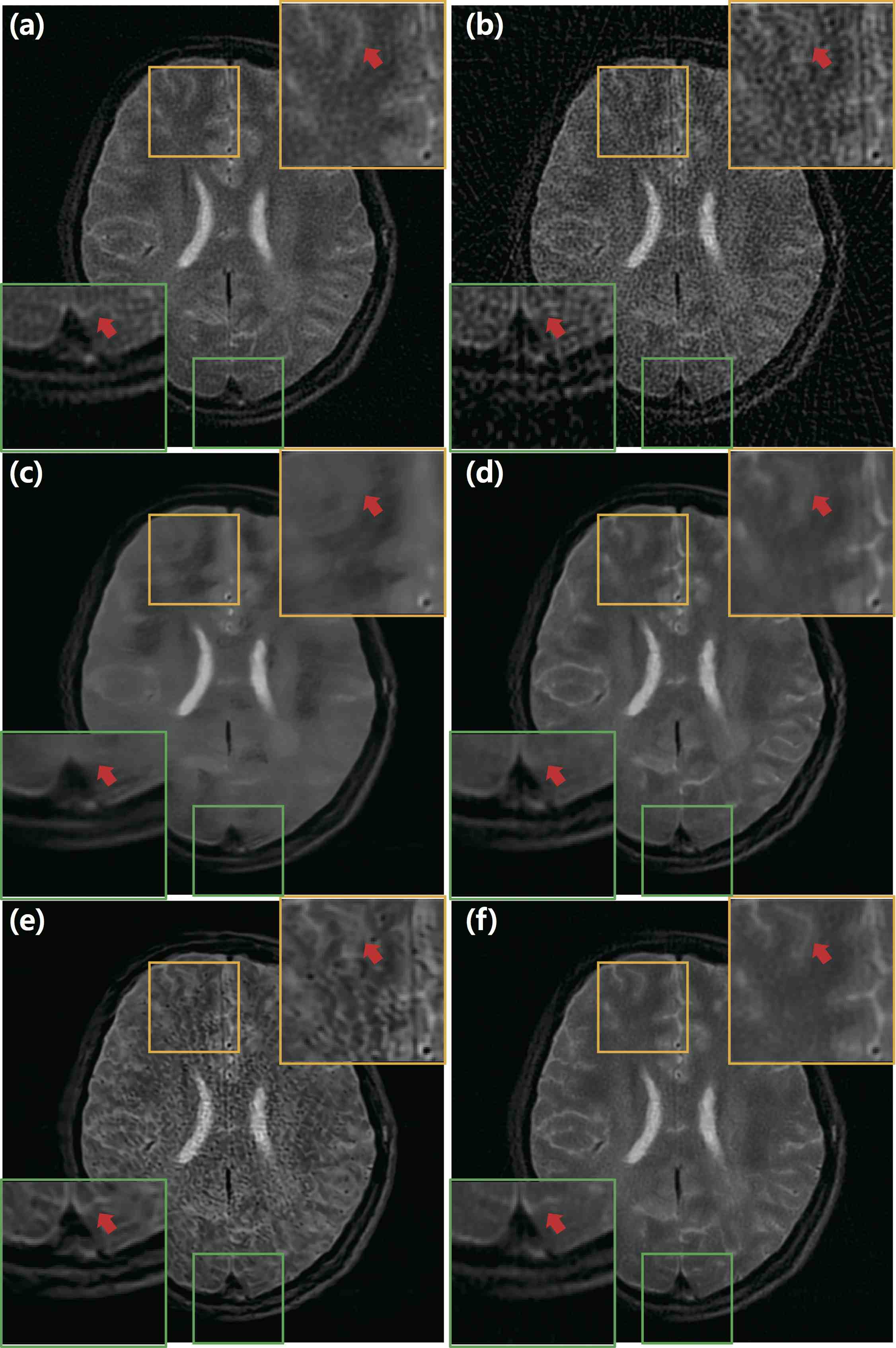}}
\caption{Effect of domain adaptation using 1 MR slice. (a) Ground truth (MR), (b) 45 view reconstruction. Reconstruction results from
CT pre-trained network (c) without fine-tuning,
and (d) with fine-tuning using 1 radial MR slice. Reconstruction results from HCP pre-trained network (e) without fine-tuning,
and (f) with fine-tuning using 1 radial MR slice. }
\label{fig:fine_tuning}
\end{figure}

Fig. \ref{fig:fine_tuning}(c) shows the brain  result from the CT pre-trained network,  when an under-sampled MR image as shown in Fig. \ref{fig:fine_tuning}(b) is used as the network input. Because the CT pre-trained network is focused on restoring CT global deficient components, the network recognized the detailed MR structures as CT global artifacts and erroneously removed them. Moreover, the pre-trained network was trained using fan-beam CT geometry instead of parallel-beam geometry, as in projection-reconstruction MR, such that the resulting global artifacts were statistically different. Therefore, the detailed structures of MR were not fully recovered, and the restored images appear as CT images. However,
domain adaptation using a fine-tuning with radial MR data significantly improves the reconstruction performance.
Specifically, Fig. \ref{fig:fine_tuning}(d) shows a restored image from the fine-tuned network with only one MR dataset. The detailed structure was better conserved in the fine-tuned network than in the pre-trained network, as compared in Fig. \ref{fig:fine_tuning}(c) and Fig. \ref{fig:fine_tuning}(d).
On the other hand,  Fig. \ref{fig:fine_tuning}(e) illustrates the baseline result from the HCP pre-trained network. The detailed MR structure was conserved, whereas the result in \ref{fig:fine_tuning}(c) from CT pre-trained network did not restore the fine detail. Moreover, with additional fine tuning with one MR slice,
 Fig. \ref{fig:fine_tuning}(f)  shows MR reconstruction images more realistic than Fig. \ref{fig:fine_tuning}(d).
The HCP pre-trained network can successfully removes global artifact, but the network preserves the detailed MR structures in contrast with the CT pre-trained network. 

\begin{figure}[!b] 	
\centering
\begin{subfigure}{\textwidth}
\centering
{\includegraphics[width=1\linewidth]{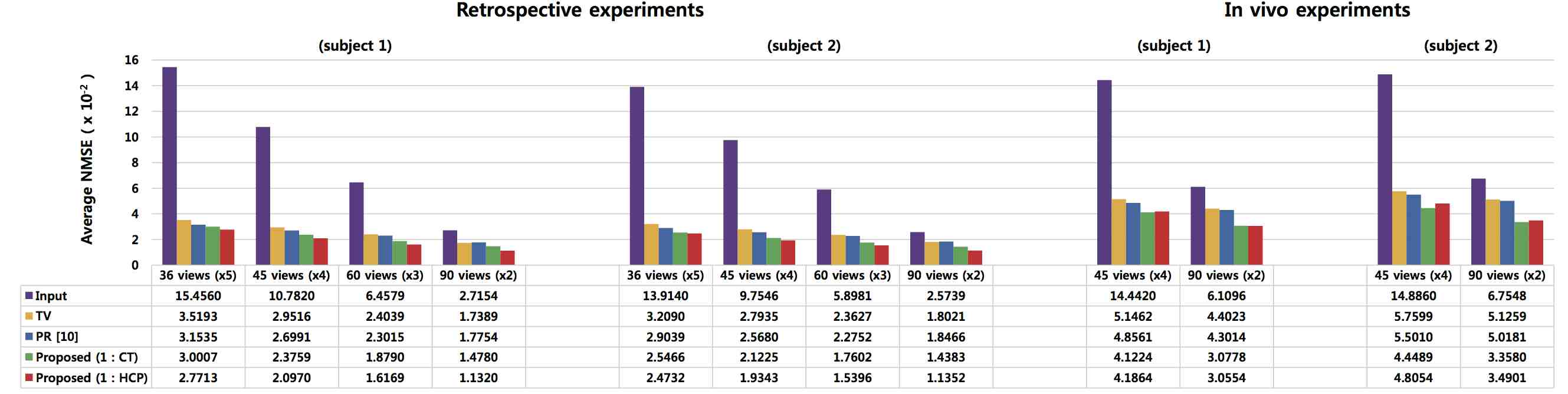}}
\caption{Average NMSE values with respect to the reconstruction methods from retrospective and in-vivo sub-sampling experiments.
Here, TV is the total variation method, PR denotes the PR-FOCUSS,  and ``Proposed (x:CT)'' and ``Proposed (x: HCP)'' refer to the fine tuning with x-MR slices when the
pretrained networks are from CT and synthesized radial MR from HCP data, respectively.}
\end{subfigure}

\vspace{0.5cm}
\centering
\begin{subfigure}{\textwidth}
\centering
{\includegraphics[width=0.6\linewidth]{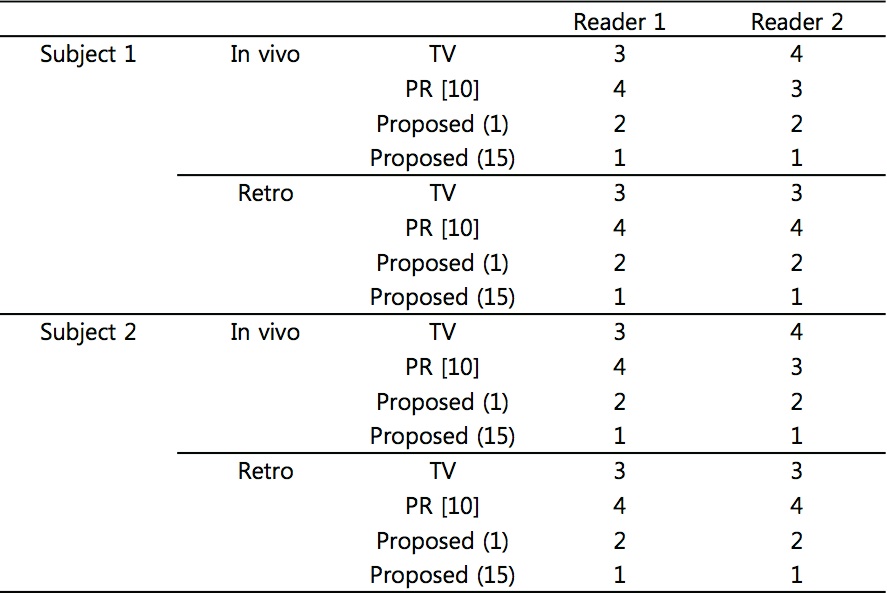}}
\caption{Radiologist blind evaluation ranking for various reconstruction methods from 90 views.}
\end{subfigure}

\vspace{0.5cm}
\centering
\begin{subfigure}{\textwidth}
\centering
{\includegraphics[width=1\linewidth]{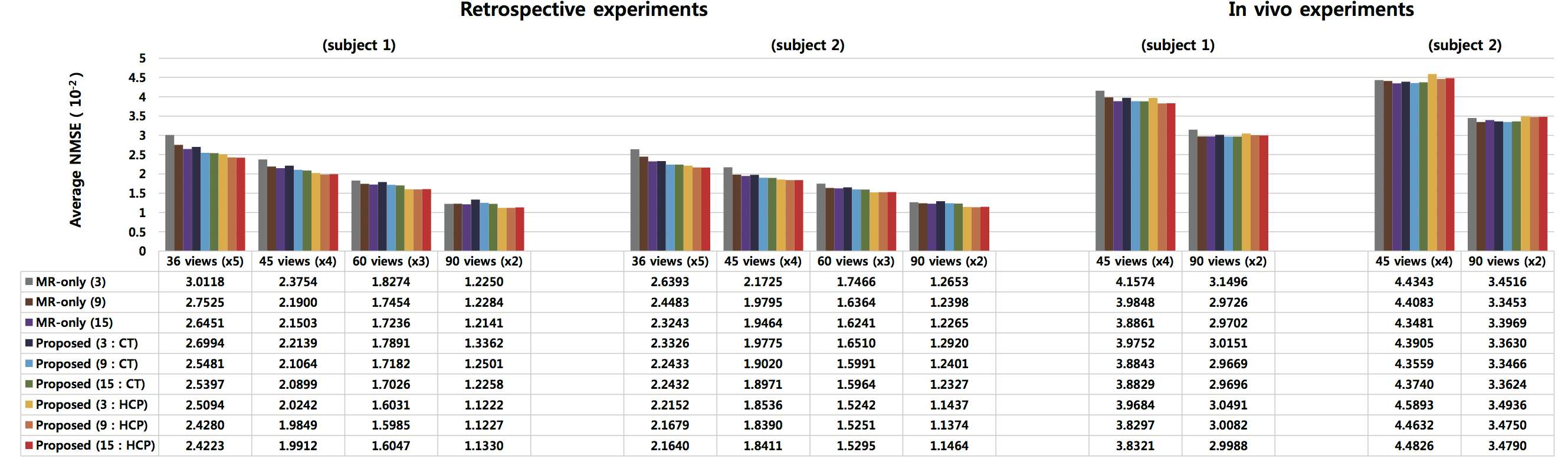}}
\caption{Average NMSE values  with respect to the training data size from retrospective and in-vivo sub-sampling experiments.
Here, TV is the total variation method,  PR denotes the PR-FOCUSS,  and ``Proposed (x:CT)'' and ``Proposed (x: HCP)'' refer to the fine tuning with x-MR slices when the
pretrained networks are from CT and synthesized radial MR from HCP data, respectively.}
\end{subfigure}

\caption{NMSE and radiologist blind evaluation results for brain dataset. }
\label{tbl:err_table}
\end{figure}

For a quantitative evaluation of the network performance with respect to the different pre-training strategy, 90, 60, 45, and 36 projection views were retrospectively sub-sampled from 180 views of projection data of two other subjects (subjects 1 and 2). Assuming that 180 views correspond to fully sampled data, Fig. \ref{tbl:err_table}(a) shows the corresponding normalized mean square error (NMSE) values. The average NMSE values were calculated by averaging twenty slice restoration results. The results clearly showed that the two proposed networks exceeded CS methods such as the TV and PR-FOCUSS\cite{ye2007projection}. Among the two domain adaptation schemes using CT and HCP pre-trained network,  the HCP pre-trained network has better performance than the CT pre-trained network. 

\begin{figure}[!b] 	
\centering
\begin{subfigure}{\textwidth}
\centering
{\includegraphics[width=0.9\linewidth]{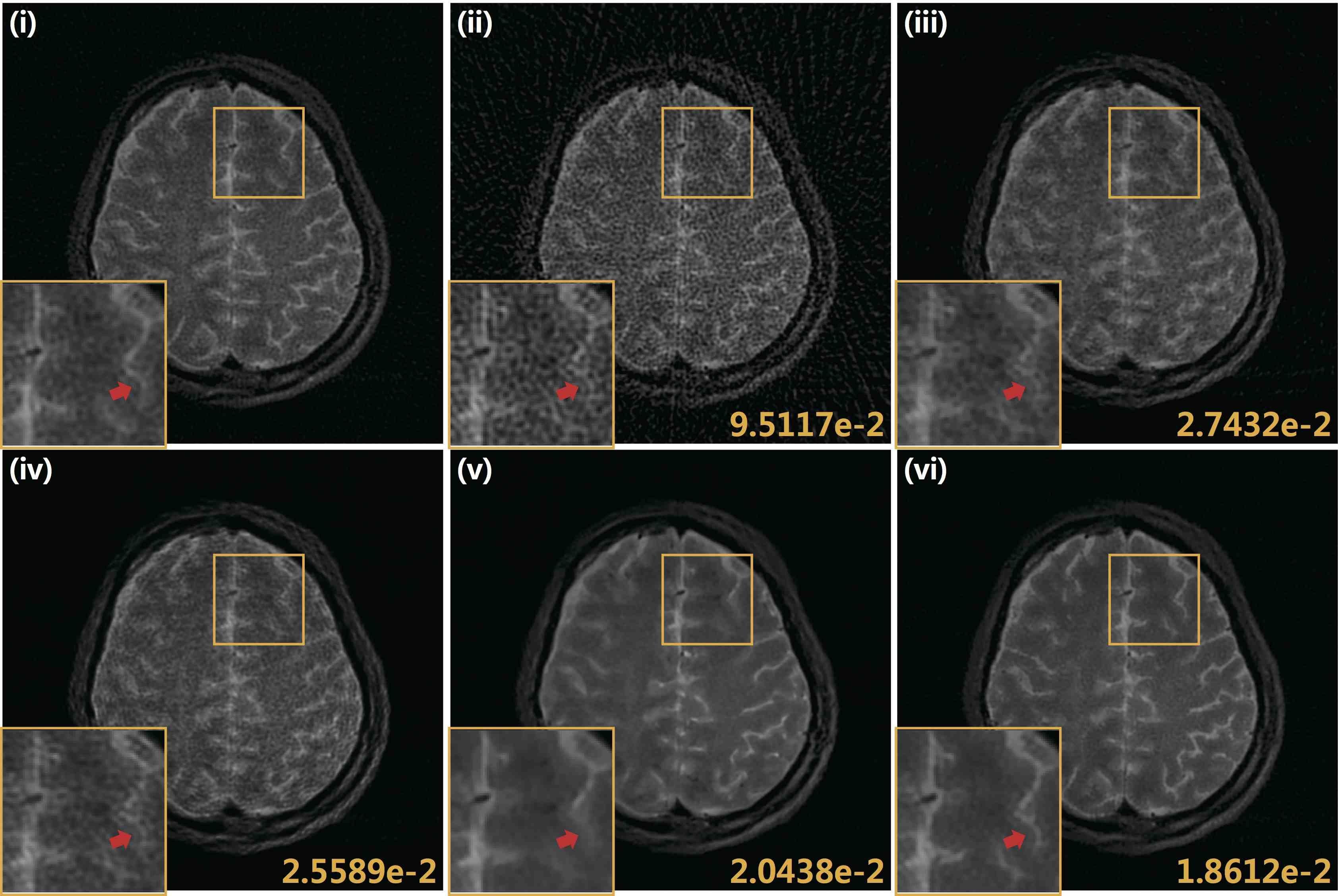}}
\caption{Reconstruction results from 45 projection views from a retrospective sub-sampling experiment.}
\end{subfigure}

\vspace{0.5cm}
\begin{subfigure}{\textwidth}
\centering
{\includegraphics[width=0.9\linewidth]{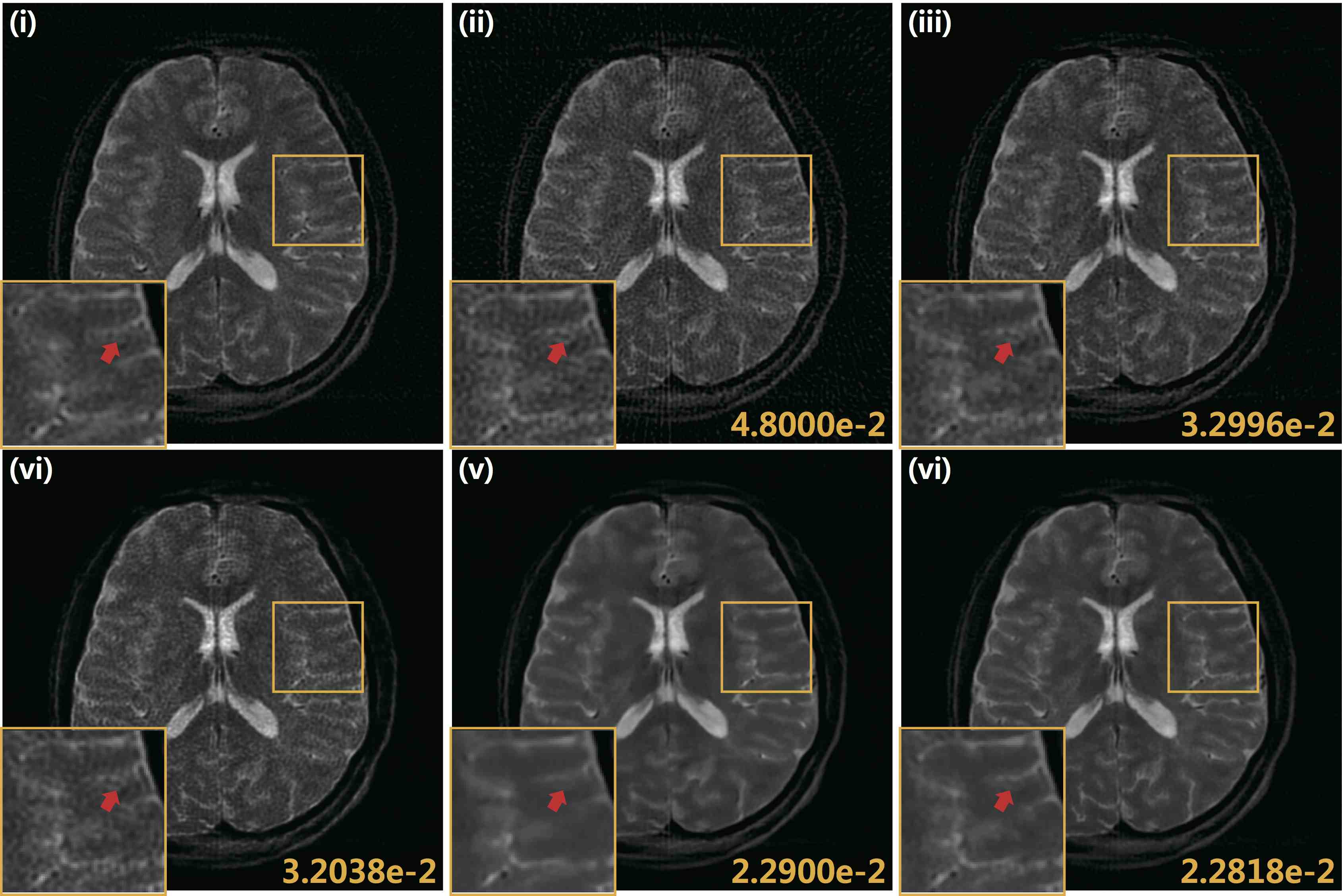}}
\caption{Reconstruction results from 90 projection views from in vivo down-sampling experiment.}
\end{subfigure}

\caption{ (i) Ground truth, (ii) X : input, (iii) TV, (iv) PR-FOCUSS, (v) domain adaptation with CT pre-trained network, (vi) domain adaptation with HCP pre-trained network.
The proposed method was fine-tuned with only 1 MR slice. The NMSE values are written at the corner.}
\label{fig:result_comp_cs1}
\end{figure}

For a qualitative comparison, Fig. \ref{fig:result_comp_cs1}(a) shows the reconstruction results from retrospective sub-sampling data with 45 views using TV and PR-FOCUSS as well as the proposed network fine-tuned with only one MR dataset from subject 0. The subjective image quality as well as the NMSE value of the slice  in Fig \ref{fig:result_comp_cs1}(a) clearly demonstrated that the proposed method restores the global artifact by eliminating streaking artifacts and preserving the detailed brain structure for all of the views.
Next, in vivo downsampled data with 90 and 45 view projections were acquired from subjects 1 and 2. Fully sampled projection data with 180 views were also acquired separately. The quantitative results in Fig \ref{tbl:err_table}(a) and the subjective image quality in Fig \ref{fig:result_comp_cs1}(b) from 90 views also confirmed that the proposed network can be generalized well to  in vivo accelerations with significantly improved image quality. In these in vivo experiments, the NMSE value for the reconstruction was calculated from 180 views acquired in a separate scan.

To evaluate the diagnostic quality of the reconstruction, two radiologists (HHK and HJS) performed a blind evaluation of the various reconstruction methods from 90  view data from subject 1 and subject 2.
As shown in Fig. \ref{tbl:err_table}(b), in both the in vivo and  retrospective downsampling cases,   the proposed method outperformed the TV and PR-FOCUSS reconstruction. Moreover, radiologists found that  the results from the proposed method with 15-slice fine-tuning did not differ from those of the fully sampled case and that the proposed method with 1-slice fine-tuning still offered very good diagnostic quality. On the other hand, the results from PR-FOCUSS and TV showed many artifacts, which affected the diagnosis.

In addition to the quality improvement, the computation time was significantly reduced using the proposed method. The proposed method required only 0.05 seconds for restoration, while the computational times for the TV and PR-FOCUSS methods were approximately 24 to 38 seconds and nearly 29 to 60 seconds, respectively.

\begin{figure}[!b] 	
\centering

\begin{subfigure}{\textwidth}
\centering
{\includegraphics[width=0.75\linewidth]{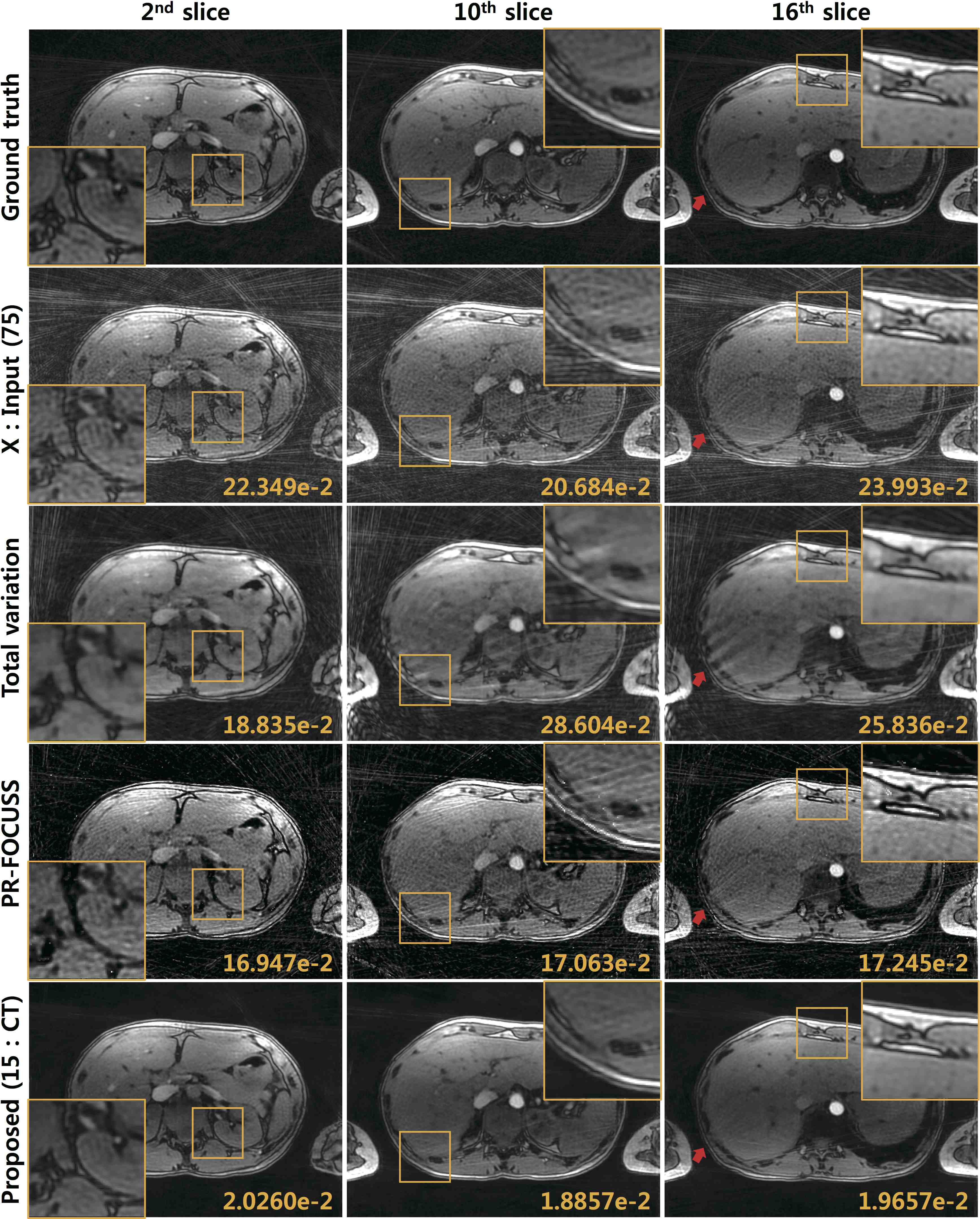}}
\caption{The abdominal reconstruction results from 75 projection views by TV, PR-FOUCSS and the proposed network fine-tuned with 15 MR-slice.
The network was pre-trained using CT data.  The NMSE values are written at the corner.}
\end{subfigure}

\vspace{0.1cm}

\begin{subfigure}{\textwidth}
\centering
{\includegraphics[width=0.75\linewidth]{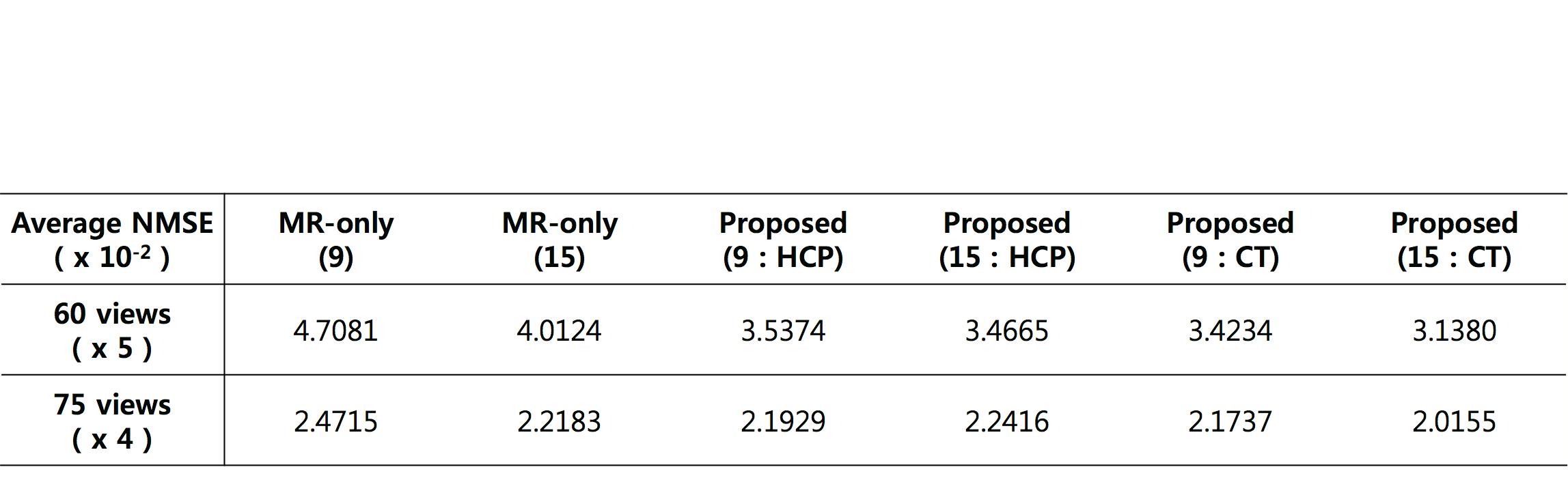}}
\caption{Average NMSE values  with respect to the training data size from retrospective sub-sampling experiments. Here, 
``MR-only (x)'' denotes the network trained with x-MR slices,  and ``Proposed (x:CT)'' and ``Proposed (x: HCP)'' refer to the fine tuning with x-MR slices when the
pretrained networks are from CT and synthesized radial MR from HCP data, respectively.}
\end{subfigure}

\caption{The abdominal reconstruction results.}
\label{fig:result_lvr_db15}
\end{figure}

Fig. \ref{fig:result_lvr_db15}(a) shows the abdomen results for the representative slices from subject 1 when the CT pre-trained network was fine-tuned network with the 15 radial MR dataset. The ground truth with 302 views was obtained by the golden-angle method and the downsampled data were generated by collecting 75 views from the first part of the golden-angle projection data. Although the detailed structure of abdominal images is more complex than that of the brain image, the proposed method  exhibited stable restoration performance. 
Fig. \ref{fig:result_lvr_db15}(b) shows a quantitative evaluation of the two different domain adaptation scheme  when 302 views projection data were retrospectively sub-sampled with 75, and 60 views projection data. 
In contrast with Fig. \ref{tbl:err_table}(c), 
 CT pre-trained network is quantitatively better than HCP pre-trained network.
 This implies that the domain adaptation from similar organ is more important than using the same imaging modality.
%
 


\section*{Discussions}\label{sec:discussion}

\subsection*{Advantages of domain adaptation}

\begin{figure}[!b] 	
\centering

{\includegraphics[width=1.0\linewidth]{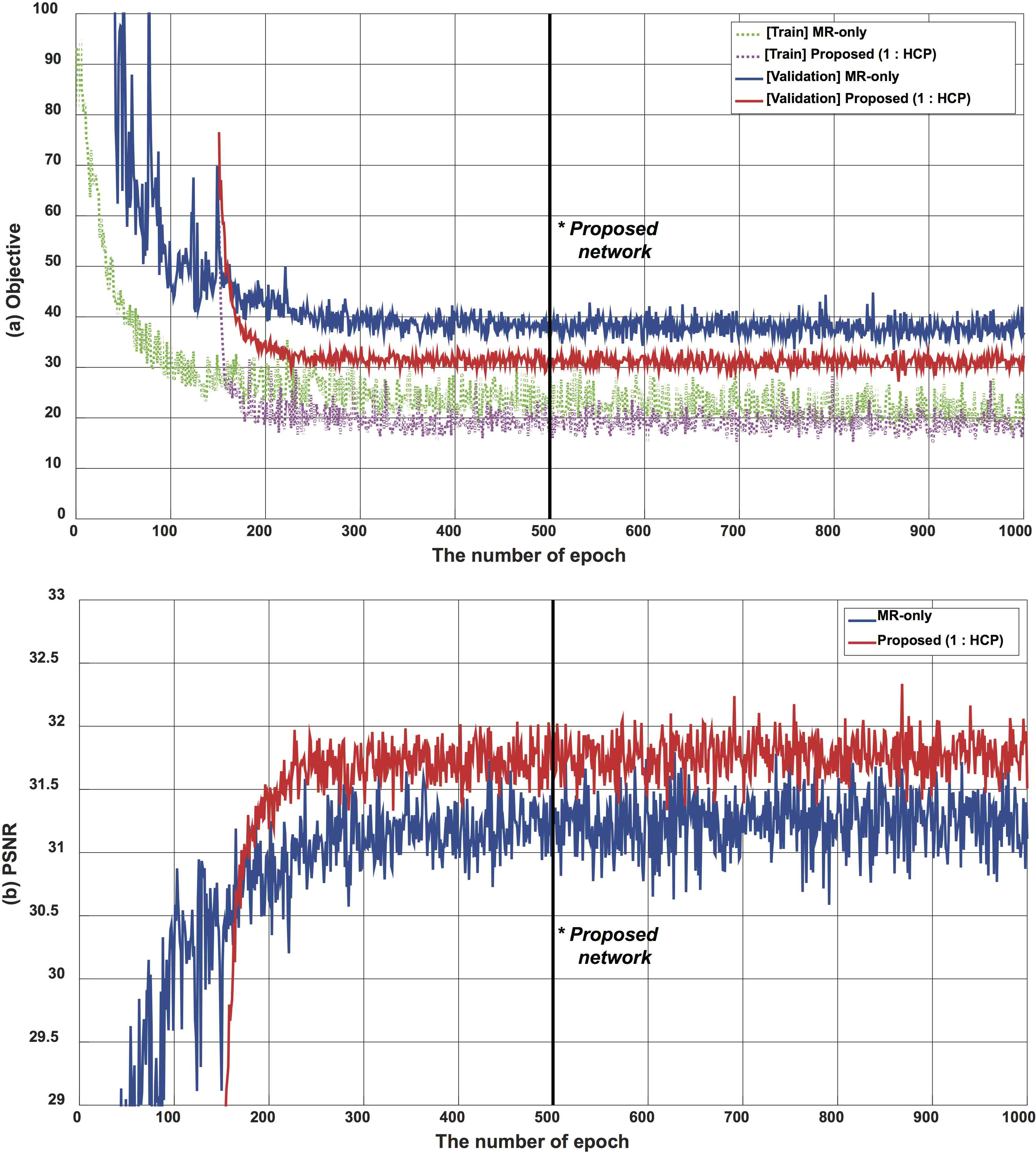}}

\caption{Convergence plots for (a) objective function and (b) peak-signal-to-noise ratio (PSNR) with respect to each epoch.}
\label{fig:err_plots}
\end{figure}

In order to verify the importance of the domain adaptation, we conducted various comparative studies. First, using the brain dataset, we compared the performance of the proposed network with that of another baseline network trained only with a few radial MR data. For a fair comparison, the network parameters for the MR-only network were equated with those of the proposed network architecture, except for the learning rate. The number of epochs used was 500, equal to that of the proposed method. 
The convergence plot is shown in Fig \ref{fig:err_plots}. The objective and PSNR values were calculated for the labels and restored images with a patch size of $256 \times 256$ when the network is trained using retrospective down-sampled images of 36, 45, 60, and 90 views. The dash-line and solid-line define objective functions for train phase and validation phase, respectively. Although the networks were trained with up to 1,000 epochs, they converged reliably after the 500th epoch without over-fitting or under-fitting. Hence, the networks for the 500th epoch were used to restore the MR images. Compared with the MR-only residual network, the domain adaptation scheme showed superior performance.

\begin{figure}[!b] 	
\centering
{\includegraphics[width=1.0\linewidth]{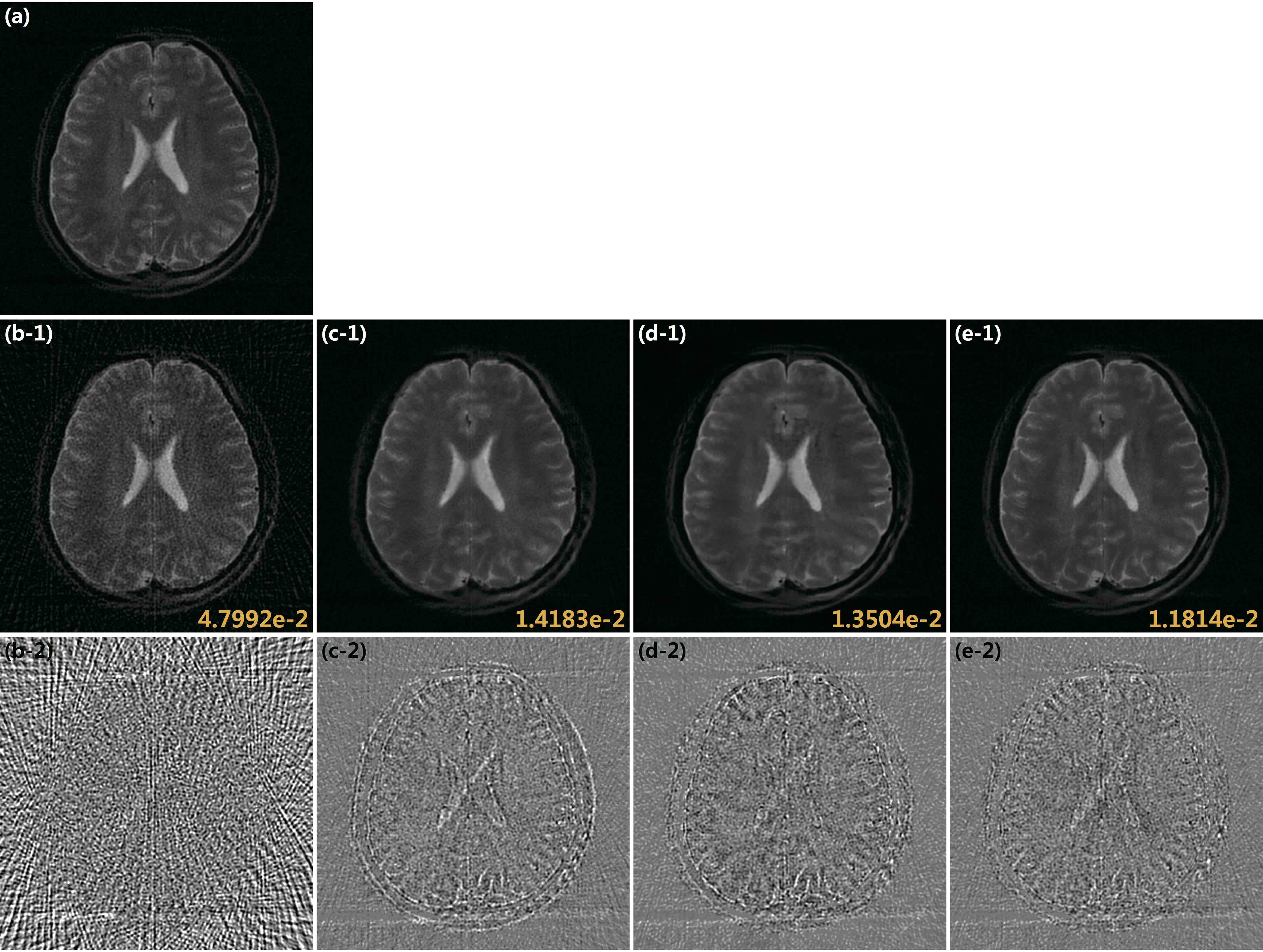}}
\caption{(a) Ground truth, (b) 60 view reconstruction, (c) MR-only network, and domain adaptation using (d) CT pre-trained network, and (e) HCP  pre-trained network. The proposed networks was fine-tuned with 1 MR-slice. Reconstructed images (*-1) and Residual images (*-2) are displayed in the second and last line, respectively. The NMSE values are written at the corner. }
\label{fig:result_comp_cs2}
\end{figure}

The restoration results from 60 view-projection data instances are shown in Fig. \ref{fig:result_comp_cs2}. Domain adaptation resulted in improved restoration, as confirmed in the difference images. Moreover,
for the case of brain data, the domain adaptation using HCP pre-training has less residual signals in the edges, indicating that the
reconstruction performance improved compared to the CT pre-trained network.
In addition, as revealed from the quantitative results in Fig. \ref{tbl:err_table}(c) for the retrospective and in vivo sub-sampling cases, 
 the proposed networks significantly outperform the MR-only residual network, as using only a few MR datasets was not enough to allow the learning of the strong global artifact pattern caused by severely under-sampled projection views. 
 

While the radial trajectory is not widely used, one area where they are increasingly used is for phase contrast flow imaging. In this case, the recovering phase information is useful. In this case, we could train the phase network separately
similar to \cite{lee2017deepres} . In this case, the magnitude network we have described so far can guide the training and inference of phase network. However, this is beyond the scope of this paper.

Readers may wonder whether we need additional data for source domain $\Sc$, such as CT dataset and synthetic MR dataset to use the proposed method. However, this is not the case, as the pre-trained network is feasible as a starting point. The situation is analogous to existing classification studies in the computer vision community using domain adaptation. For example,  AlexNet \cite{krizhevsky2012imagenet} is  a classical pre-trained network for the ImageNet dataset \cite{deng2009imagenet}. If one is interested in designing an image classification network for a microscopy dataset, for instance, it would be feasible to start with AlexNet and fine-tune it with a microscopy dataset.
The designer does not need to run additional pre-training for the ImageNet dataset; therefore, the only additional complexity and datasets come from the fine-tuning stage. Similarly, in our MR network, as a pre-trained network is used as a starting point, the proposed network is quickly trained and converges to a good solution, even with a small amount of MR data. This is an important advantage of the proposed network compared to the existing deep learning approach for MR, which requires a considerably large dataset. In fact, our pre-trained network  will be publicly available on the authors' homepage upon publication so that readers can use it for fine-tuning.

One may be concerned about whether the fine-tune network can deteriorate the pre-trained network. 
However, the good news is that users should not be concerned about the effect on the pre-trained network from the fine-tuning process because if the user is interested in the CT reconstruction problem, the original pre-trained network can be used for their purpose. Note that our goal is not to design a network that is optimal for both CT and MR data. Rather, our goal of domain adaptation is to design a network that is optimal for MR data. 

\subsection*{Dependency on the fine-tuning data size}

\begin{figure}[!b] 	
\centering
\begin{subfigure}{\textwidth}
\centering
{\includegraphics[width=1.0\linewidth]{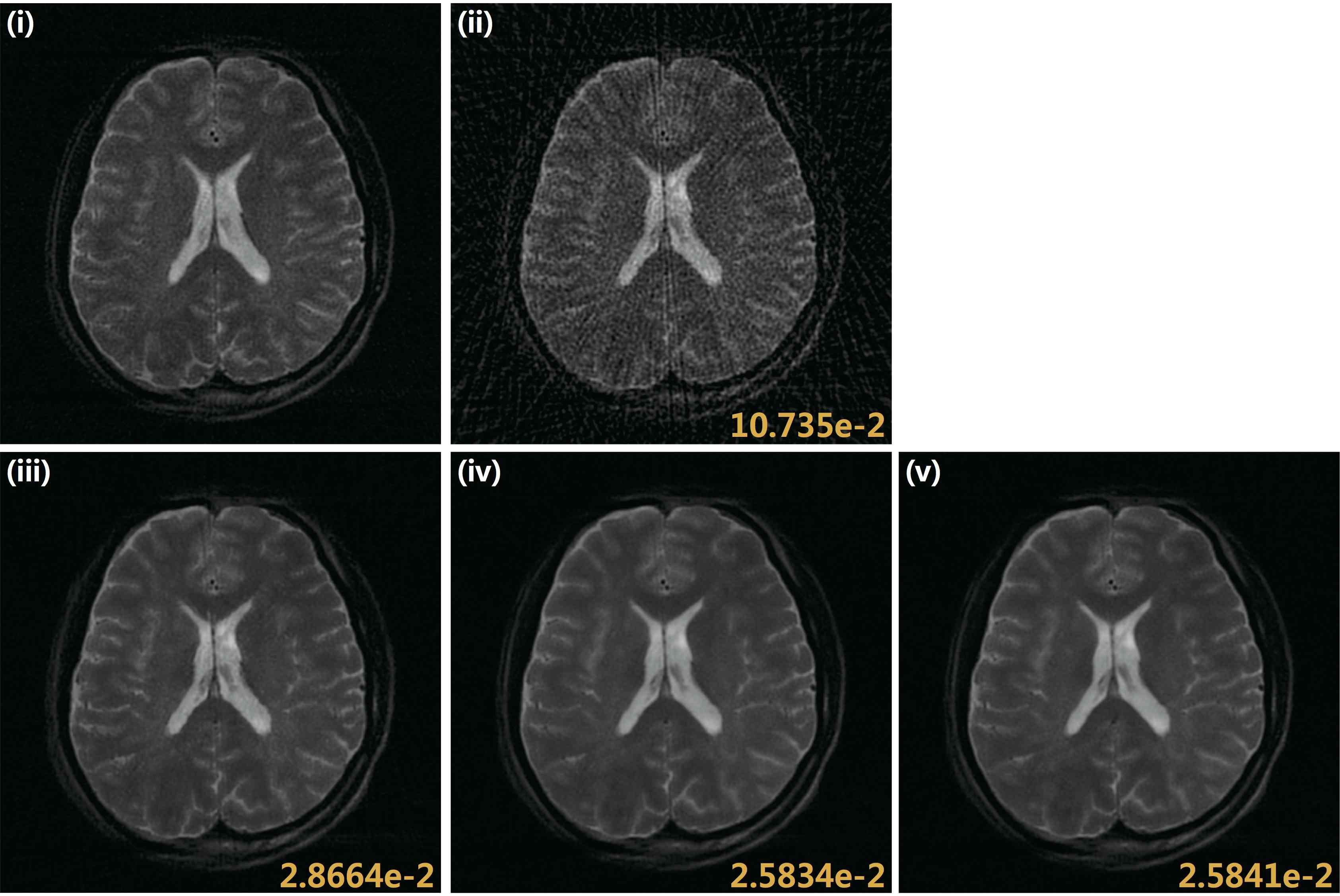}}
\caption{Brain reconstruction results from 45 projection views from in vivo down-sampling experiment.}
\end{subfigure}

\vspace{0.5cm}
\centering
\begin{subfigure}{\textwidth}
\centering
{\includegraphics[width=1.0\linewidth]{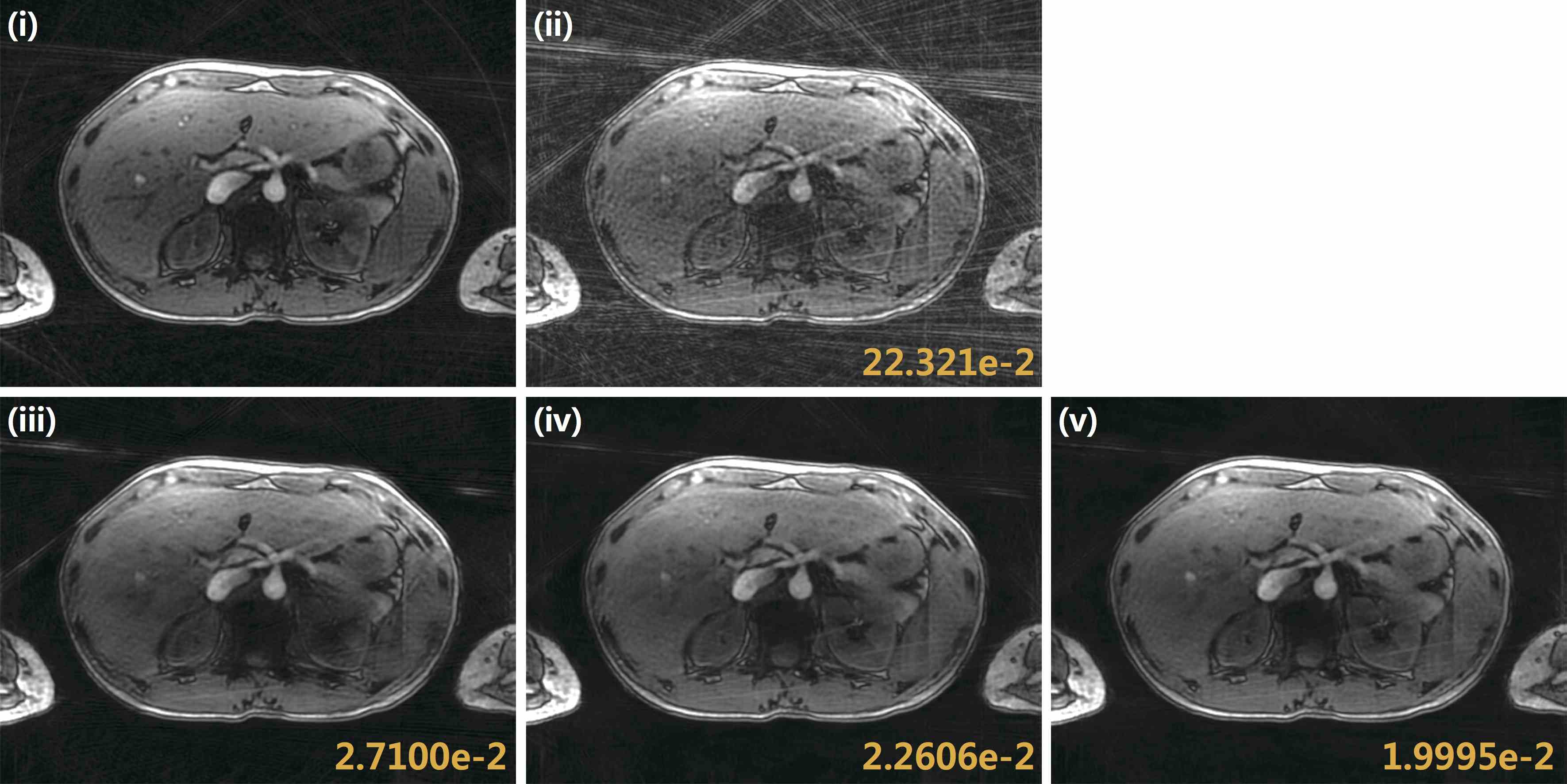}}
\caption{Abdomen reconstruction results from 60 projection views from in vivo down-sampling experiment.}
\end{subfigure}

\caption{The brain and abdominal reconstruction results from in vivo 45 and 75 projection views, respectively. (i) Ground truth, (ii) FBP reconstruction using down-sampled projection view, and the proposed network
finely tuned with 
(iii) 1,  (iv) 9 and  (v)15 MR slices, respectively. The NMSE values are written at the corner.}
\label{fig:result_comp_cs3}
\end{figure}

In order to investigate the dependency on the fine-tuning dataset, we fine-tuned the network with a various numbers of MR datasets. Figs. \ref{fig:result_comp_cs3}(a) and (b) show in vivo experimental results for brain and abdominal
data set with various size of fine tuning data set.  Using more radial MR data  improved the quality of the restored images. This subjective image quality coincides with the quantitative results shown in Fig. \ref{tbl:err_table}(c), showing reduced average NMSE values with more radial  MR datasets for fine-tuning. 
Note that the abdominal image has more complex structures when compared with the brain image. Owing to the image details, the fine-tuning steps for the abdominal image require more MR datasets and more epochs, and we found that the 15 MR dataset and 3,000 epochs are reasonable. Again, with this small amount of data, it is remarkable to observe the superior image reconstruction quality  in Fig. \ref{fig:result_lvr_db15} relative to the outcomes of all existing methods.

In addition, as shown in Fig \ref{tbl:err_table}(c), an increase in the number of MR datasets improves the performance of the MR-only network, and it is impressive to note that the MR-only residual network trained with more than three MR slices outperforms the CS-based approaches (see Figs. \ref{tbl:err_table}(a) and \ref{tbl:err_table}(b)). This again demonstrates the advantages of the proposed learning architecture. While the MR-only residual network shows a marked improvement with the number of MR datasets, the proposed domain-adaptation method is capable of more stable reconstruction than the MR-only network. 

\subsection*{Acceleration Ratio}

The current method for the brain data and abdomen dataset requires at least  36 and 75 projection views, respectively. Some readers may wonder whether this is a significant acceleration factor considering the latest radial CS works \cite{feng2014golden,cruz2015accelerated}, which require far fewer radial spokes. However, it is important to note that those works \cite{feng2014golden,cruz2015accelerated} are {\em dynamic} CS works that exploit temporal redundancies, whereas our work is for {\em static} MR reconstruction, which only exploits the redundancies within the frame rather than adjacent temporal frames. In the aforementioned dynamic contrast-enhancement MRI studies \cite{feng2014golden,cruz2015accelerated}, the structures remain the same except for breathing-related motions. Therefore, there are many temporal redundancies that can be exploited to achieve this higher acceleration factor. However, to the best of our knowledge, we are not aware of any {\em static} CS MR works that can achieve such a high acceleration factor in actual datasets such as those from the brain and liver.

In fact, an extension of the proposed method  for dynamic MR images may be feasible and a significantly higher acceleration factor may be achievable using additional temporal redundancy. This is a very important topic in its own right, and it will be reported elsewhere.

\subsection*{Generalization performance}

As discussed earlier, the higher the acceleration factor is, the fewer the radial k-space sampling trajectories are in the Fourier domain, which leads to more severe streaking artifacts. However, one of the most important advantages of the proposed deep learning approach is its generalization power. More specifically, streaking artifacts during the training phase do not need to be identical in the test phase. For example, our CT network is trained using a fan-beam trajectory with different acceleration factors, whereas projection-reconstruction MR is a type of parallel-beam geometry.  However, even without applying domain adaptation, most of the streaking artifacts can be removed (see Fig.~\ref{fig:fine_tuning}). Such remarkable generalization power of a deep network has been an intensive research interest in the statistical learning theory community, and many studies attribute this to the exponential representation power  of a deep network \cite{telgarsky2016benefits}. Therefore, as long as we can avoid over-fitting, the same network can be generalized well for various acceleration factors. Interestingly, to avoid over-fitting, we found that training with various acceleration factors was helpful, as described earlier. More specifically, when the proposed network was trained using datasets from  various acceleration factors, it was better trained with little over-fitting and was able to reconstruct corrupted images at various acceleration factors.

\section*{Conclusion}\label{sec:conclusion}

We developed a novel deep learning approach with domain-adaptation to reconstruct high quality images from sub-sampled k-space data in MRI. The proposed network employed  a pre-trained network using  CT datasets or synthetic
radial MR data, with fine-tuning using a small number of radial  MR dataset. If there is a sufficient amount of radial MR data, the deep network with MR data alone showed good restoration results. However, for an insufficient training dataset, the proposed network showed the best performance when combined with pre-training and fine-tuning steps.
Among the various factors such as organ and imaging modality, our experiments showed that the
similar organ structure is more important, so it is better to use a pre-trained network from similar organ structures.
 Moreover, the computation speed was much faster than the speeds of conventional compressed sensing methods because the proposed method does not use projection or back-projection, both of which require heavy computational complexity.

The proposed domain-adaptation approach demonstrated the potential for mixing different medical systems when the image artifacts are similar and topologically simple. The principles of medical imaging systems such as computed tomography (CT), magnetic resonance image (MRI), and optical diffraction tomography (ODT) are closely related in the Fourier space according to the Fourier slice theorem and the Fourier diffraction theorem. Based on this relationship, the domain-adaptation approach can be applied to various modalities to remove various artifacts.

\section*{Acknowledgment}

The authors would like to thanks Dr. Cynthia MaCollough, the Mayo Clinic, the American Association of Physicists in Medicine (AAPM), and grant EB01705 and EB01785 from the National Institute of Biomedical Imaging and Bioengineering for providing the Low-Dose CT Grand Challenge data set. This work is supported by National Research Foundation of Korea, Grant number NRF-2016R1A2B3008104 and NRF-2013M3A9B2076548.
The data for a pre-trained network were provided by the Human Connectome Project, MGH-USC Consortium. Thus, this work acknowledges funding from the National Institutes of Health, NIH Blueprint Initiative for Neuroscience Research grant U01MH093765, National Institutes of Health grant P41EB015896, and NIH NIBIB grant K99/R00EB012107


\begin{thebibliography}{10}
\providecommand{\url}[1]{#1}
\csname url@samestyle\endcsname
\providecommand{\newblock}{\relax}
\providecommand{\bibinfo}[2]{#2}
\providecommand{\BIBentrySTDinterwordspacing}{\spaceskip=0pt\relax}
\providecommand{\BIBentryALTinterwordstretchfactor}{4}
\providecommand{\BIBentryALTinterwordspacing}{\spaceskip=\fontdimen2\font plus
\BIBentryALTinterwordstretchfactor\fontdimen3\font minus
  \fontdimen4\font\relax}
\providecommand{\BIBforeignlanguage}[2]{{%
\expandafter\ifx\csname l@#1\endcsname\relax
\typeout{** WARNING: IEEEtran.bst: No hyphenation pattern has been}%
\typeout{** loaded for the language `#1'. Using the pattern for}%
\typeout{** the default language instead.}%
\else
\language=\csname l@#1\endcsname
\fi
#2}}
\providecommand{\BIBdecl}{\relax}
\BIBdecl

\bibitem{bernstein2004handbook}
M.~A. Bernstein, K.~F. King, and X.~J. Zhou, \emph{Handbook of {MRI} pulse
  sequences}.\hskip 1em plus 0.5em minus 0.4em\relax Elsevier, 2004.

\bibitem{jung1991reduction}
K.~Jung and Z.~Cho, ``Reduction of flow artifacts in {NMR} diffusion imaging
  using view-angle tilted line-integral projection reconstruction,''
  \emph{Magnetic Resonance in Medicine}, vol.~19, no.~2, pp. 349--360, 1991.

\bibitem{gmitro1993use}
A.~F. Gmitro and A.~L. Alexander, ``Use of a projection reconstruction method
  to decrease motion sensitivity in diffusion-weighted {MRI},'' \emph{Magnetic
  Resonance in Medicine}, vol.~29, no.~6, pp. 835--838, 1993.

\bibitem{katoh2006mr}
M.~Katoh, E.~Spuentrup, A.~Buecker, W.~J. Manning, R.~W. G{\"u}nther, and R.~M.
  Botnar, ``{MR coronary vessel wall imaging: Comparison between radial and
  spiral k-space sampling},'' \emph{Journal of Magnetic Resonance Imaging},
  vol.~23, no.~5, pp. 757--762, 2006.

\bibitem{trouard1996analysis}
T.~P. Trouard, Y.~Sabharwal, M.~I. Altbach, and A.~F. Gmitro, ``Analysis and
  comparison of motion-correction techniques in diffusion-weighted imaging,''
  \emph{Journal of Magnetic Resonance Imaging}, vol.~6, no.~6, pp. 925--935,
  1996.

\bibitem{donoho2006compressed}
D.~L. Donoho, ``Compressed sensing,'' \emph{IEEE Transactions on Information
  Theory}, vol.~52, no.~4, pp. 1289--1306, 2006.

\bibitem{candes2006robust}
E.~J. Cand{\`e}s, J.~Romberg, and T.~Tao, ``Robust uncertainty principles:
  Exact signal reconstruction from highly incomplete frequency information,''
  \emph{IEEE Transactions on Information Theory}, vol.~52, no.~2, pp. 489--509,
  2006.

\bibitem{lustig2007sparse}
M.~Lustig, D.~Donoho, and J.~M. Pauly, ``{Sparse MRI: The application of
  compressed sensing for rapid MR imaging},'' \emph{Magnetic resonance in
  medicine}, vol.~58, no.~6, pp. 1182--1195, 2007.

\bibitem{jung2009k}
H.~Jung, K.~Sung, K.~S. Nayak, E.~Y. Kim, and J.~C. Ye, ``k-t focuss: A general
  compressed sensing framework for high resolution dynamic mri,''
  \emph{Magnetic Resonance in Medicine}, vol.~61, no.~1, pp. 103--116, 2009.

\bibitem{ye2007projection}
J.~C. Ye, S.~Tak, Y.~Han, and H.~W. Park, ``{Projection reconstruction MR
  imaging using FOCUSS},'' \emph{Magnetic Resonance in Medicine}, vol.~57,
  no.~4, pp. 764--775, 2007.

\bibitem{jung2010radial}
H.~Jung, J.~Park, J.~Yoo, and J.~C. Ye, ``Radial k-t focuss for high-resolution
  cardiac cine mri,'' \emph{Magnetic Resonance in Medicine}, vol.~63, no.~1,
  pp. 68--78, 2010.

\bibitem{feng2014golden}
L.~Feng, R.~Grimm, K.~T. Block, H.~Chandarana, S.~Kim, J.~Xu, L.~Axel, D.~K.
  Sodickson, and R.~Otazo, ``{Golden-angle radial sparse parallel MRI:
  Combination of compressed sensing, parallel imaging, and golden-angle radial
  sampling for fast and flexible dynamic volumetric MRI},'' \emph{Magnetic
  resonance in medicine}, vol.~72, no.~3, pp. 707--717, 2014.

\bibitem{cruz2015accelerated}
G.~Cruz, D.~Atkinson, C.~Buerger, T.~Schaeffter, and C.~Prieto, ``{Accelerated
  motion corrected three-dimensional abdominal MRI using total variation
  regularized SENSE reconstruction},'' \emph{Magnetic resonance in medicine},
  2015.

\bibitem{krizhevsky2012imagenet}
A.~Krizhevsky, I.~Sutskever, and G.~E. Hinton, ``Imagenet classification with
  deep convolutional neural networks,'' in \emph{Advances in Neural Information
  Processing Systems}, 2012, pp. 1097--1105.

\bibitem{ronneberger2015u}
O.~Ronneberger, P.~Fischer, and T.~Brox, ``U-net: Convolutional networks for
  biomedical image segmentation,'' in \emph{International Conference on Medical
  Image Computing and Computer-Assisted Intervention}.\hskip 1em plus 0.5em
  minus 0.4em\relax Springer, 2015, pp. 234--241.

\bibitem{zhang2016beyond}
K.~Zhang, W.~Zuo, Y.~Chen, D.~Meng, and L.~Zhang, ``Beyond a {G}aussian
  denoiser: Residual learning of deep {CNN} for image denoising,'' \emph{arXiv
  preprint arXiv:1608.03981}, 2016.

\bibitem{kim2015accurate}
J.~Kim, J.~K. Lee, and K.~M. Lee, ``Accurate image super-resolution using very
  deep convolutional networks,'' \emph{arXiv preprint arXiv:1511.04587}, 2015.

\bibitem{shi2016real}
W.~Shi, J.~Caballero, F.~Husz{\'a}r, J.~Totz, A.~P. Aitken, R.~Bishop,
  D.~Rueckert, and Z.~Wang, ``Real-time single image and video super-resolution
  using an efficient sub-pixel convolutional neural network,'' in
  \emph{Proceedings of the IEEE Conference on Computer Vision and Pattern
  Recognition}, 2016, pp. 1874--1883.

\bibitem{wang2016accelerating}
S.~Wang, Z.~Su, L.~Ying, X.~Peng, S.~Zhu, F.~Liang, D.~Feng, and D.~Liang,
  ``Accelerating magnetic resonance imaging via deep learning,'' in \emph{2016
  IEEE 13th International Symposium on Biomedical Imaging (ISBI)}.\hskip 1em
  plus 0.5em minus 0.4em\relax IEEE, 2016, pp. 514--517.

\bibitem{hammernik2016learning}
K.~Hammernik, F.~Knoll, D.~Sodickson, and T.~Pock, ``Learning a variational
  model for compressed sensing {MRI} reconstruction,'' in \emph{Proceedings of
  the International Society of Magnetic Resonance in Medicine (ISMRM)}, 2016.

\bibitem{kwon2016learning}
K.~Kwon, D.~Kim, and H.~Park, ``A parallel mr imaging method using multilayer
  perceptron,'' \emph{Medical physics}, 2017.

\bibitem{kang2016deep}
E.~Kang, J.~Min, and J.~C. Ye, ``A deep convolutional neural network using
  directional wavelets for low-dose {X-ray CT} reconstruction,'' \emph{arXiv
  preprint arXiv:1610.09736}, 2016.

\bibitem{chen2015learning}
Y.~Chen, W.~Yu, and T.~Pock, ``On learning optimized reaction diffusion
  processes for effective image restoration,'' in \emph{Proceedings of the IEEE
  Conference on Computer Vision and Pattern Recognition}, 2015, pp. 5261--5269.

\bibitem{mao2016image}
X.-J. Mao, C.~Shen, and Y.-B. Yang, ``Image denoising using very deep fully
  convolutional encoder-decoder networks with symmetric skip connections,''
  \emph{arXiv preprint arXiv:1603.09056}, 2016.

\bibitem{xie2012image}
J.~Xie, L.~Xu, and E.~Chen, ``Image denoising and inpainting with deep neural
  networks,'' in \emph{Advances in Neural Information Processing Systems},
  2012, pp. 341--349.

\bibitem{jin2016deep}
K.~H. Jin, M.~T. McCann, E.~Froustey, and M.~Unser, ``Deep convolutional neural
  network for inverse problems in imaging,'' \emph{arXiv preprint
  arXiv:1611.03679}, 2016.

\bibitem{han2016deep}
Y.~Han, J.~Yoo, and J.~C. Ye, ``Deep residual learning for compressed sensing
  {CT} reconstruction via persistent homology analysis,'' \emph{arXiv preprint
  arXiv:1611.06391}, 2016.

\bibitem{hsieh2009computed}
J.~Hsieh, ``Computed tomography: principles, design, artifacts, and recent
  advances.''\hskip 1em plus 0.5em minus 0.4em\relax SPIE Bellingham, WA, 2009.

\bibitem{yosinski2014transferable}
J.~Yosinski, J.~Clune, Y.~Bengio, and H.~Lipson, ``How transferable are
  features in deep neural networks?'' in \emph{Advances in Neural Information
  Processing Systems}, 2014, pp. 3320--3328.

\bibitem{pan2010survey}
S.~J. Pan and Q.~Yang, ``A survey on transfer learning,'' \emph{IEEE
  Transactions on Knowledge and Data Engineering}, vol.~22, no.~10, pp.
  1345--1359, 2010.

\bibitem{ben2007analysis}
S.~Ben-David, J.~Blitzer, K.~Crammer, F.~Pereira \emph{et~al.}, ``Analysis of
  representations for domain adaptation,'' \emph{Advances in Neural Information
  Processing Systems}, vol.~19, p. 137, 2007.

\bibitem{ben2010theory}
S.~Ben-David, J.~Blitzer, K.~Crammer, A.~Kulesza, F.~Pereira, and J.~W.
  Vaughan, ``A theory of learning from different domains,'' \emph{Machine
  Learning}, vol.~79, no. 1-2, pp. 151--175, 2010.

\bibitem{anthony2009neural}
M.~Anthony and P.~L. Bartlett, \emph{Neural network learning: Theoretical
  foundations}.\hskip 1em plus 0.5em minus 0.4em\relax Cambridge University
  Press, 2009.

\bibitem{bartlett2002rademacher}
P.~L. Bartlett and S.~Mendelson, ``Rademacher and gaussian complexities: Risk
  bounds and structural results,'' \emph{Journal of Machine Learning Research},
  vol.~3, no. Nov, pp. 463--482, 2002.

\bibitem{long2015fully}
J.~Long, E.~Shelhamer, and T.~Darrell, ``Fully convolutional networks for
  semantic segmentation,'' in \emph{Proceedings of the IEEE Conference on
  Computer Vision and Pattern Recognition}, 2015, pp. 3431--3440.

\bibitem{ganin2015domain}
Y.~Ganin, E.~Ustinova, H.~Ajakan, P.~Germain, H.~Larochelle, F.~Laviolette,
  M.~Marchand, and V.~Lempitsky, ``Domain-adversarial training of neural
  networks,'' \emph{arXiv preprint arXiv:1505.07818}, 2015.

\bibitem{ioffe2015batch}
S.~Ioffe and C.~Szegedy, ``Batch normalization: Accelerating deep network
  training by reducing internal covariate shift,'' \emph{arXiv preprint
  arXiv:1502.03167}, 2015.

\bibitem{he2016deep}
K.~He, X.~Zhang, S.~Ren, and J.~Sun, ``Deep residual learning for image
  recognition,'' in \emph{Proceedings of the IEEE conference on computer vision
  and pattern recognition}, 2016, pp. 770--778.

\bibitem{he2016identity}
------, ``Identity mappings in deep residual networks,'' in \emph{European
  Conference on Computer Vision}.\hskip 1em plus 0.5em minus 0.4em\relax
  Springer, 2016, pp. 630--645.

\bibitem{peters2000undersampled}
D.~C. Peters, F.~R. Korosec, T.~M. Grist, W.~F. Block, J.~E. Holden, K.~K.
  Vigen, and C.~A. Mistretta, ``Undersampled projection reconstruction applied
  to {MR} angiography,'' \emph{Magnetic Resonance in Medicine}, vol.~43, no.~1,
  pp. 91--101, 2000.

\bibitem{flohr2005image}
T.~Flohr, K.~Stierstorfer, S.~Ulzheimer, H.~Bruder, A.~Primak, and C.~H.
  McCollough, ``Image reconstruction and image quality evaluation for a
  64-slice {CT} scanner with z-flying focal spot,'' \emph{Medical physics},
  vol.~32, no.~8, pp. 2536--2547, 2005.

\bibitem{kachelriess2006flying}
M.~Kachelrie{\ss}, M.~Knaup, C.~Pen{\ss}el, and W.~A. Kalender, ``{Flying focal
  spot (FFS) in cone-beam CT},'' \emph{IEEE Transactions on Nuclear Science},
  vol.~53, no.~3, pp. 1238--1247, 2006.

\bibitem{vedaldi2015matconvnet}
A.~Vedaldi and K.~Lenc, ``Matconvnet: Convolutional neural networks for
  matlab,'' in \emph{Proceedings of the 23rd ACM international conference on
  Multimedia}.\hskip 1em plus 0.5em minus 0.4em\relax ACM, 2015, pp. 689--692.

\bibitem{goldstein2009split}
T.~Goldstein and S.~Osher, ``{The split Bregman method for L1-regularized
  problems},'' \emph{SIAM journal on imaging sciences}, vol.~2, no.~2, pp.
  323--343, 2009.

\bibitem{deng2009imagenet}
J.~Deng, W.~Dong, R.~Socher, L.-J. Li, K.~Li, and L.~Fei-Fei, ``Imagenet: A
  large-scale hierarchical image database,'' in \emph{Computer Vision and
  Pattern Recognition, 2009. CVPR 2009. IEEE Conference on}.\hskip 1em plus
  0.5em minus 0.4em\relax IEEE, 2009, pp. 248--255.

\bibitem{telgarsky2016benefits}
M.~Telgarsky, ``Benefits of depth in neural networks,'' \emph{JMLR: Workshop
  and Conference Proceedings}, vol.~49, pp. 1--23, 2016.
  
  \bibitem{lee2017deepres}
D.~Lee,  J.~Yoo, and J.C.~Ye, ``Deep residual learning for compressed sensing MRI'', in \emph{Proceedings of the IEEE Symposium on
Biomedical Imaging}, 2017, pp. 15--18.
\end{thebibliography}

\clearpage


\clearpage
%
%
%
\clearpage

\clearpage
%
%
%
%
%
%
%
%
%
%
%
%
%
%
%
%
%
%
%
%
%
%
%
\clearpage

\end{document}